\documentclass{article} 
\usepackage{arxiv}

\usepackage[utf8]{inputenc} 
\usepackage[T1]{fontenc}    

\usepackage{url}            
\usepackage{booktabs}       
\usepackage{amsfonts}       
\usepackage{nicefrac}       
\usepackage{microtype}      
\usepackage{lipsum}
\usepackage{graphicx}

\usepackage{amsmath,amssymb,amsfonts,mathtools} 
\usepackage{amsthm} %
\usepackage{mathrsfs} %
\usepackage[figuresright]{rotating} %
\usepackage[title]{appendix} %

\usepackage{textcomp} %
\usepackage{manyfoot} %
\usepackage{algorithm} %
\usepackage{algorithmicx} %
\usepackage{algpseudocode} %
\usepackage{program} %
\usepackage{listings} %

\usepackage{float} %
\usepackage{multirow} %
\usepackage{array} %
\usepackage{color} %
\usepackage{esvect} %
\usepackage[nodisplayskipstretch]{setspace} %
\usepackage{subcaption} %

\usepackage[numbers]{natbib}

\makeatletter
\def\thm@space@setup{%
  \thm@preskip=12pt plus 2pt minus 1pt
  \thm@postskip=12pt plus 2pt minus 1pt
}

\newtheoremstyle{thmstyleone} 
{18pt plus 2pt minus 1pt} 
{18pt plus 2pt minus 1pt} 
{\itshape} 
{0pt} 
{\bfseries} 
{.} 
{.5em} 
{} 

\newtheoremstyle{thmstyletwo} 
{18pt plus 2pt minus 1pt} 
{18pt plus 2pt minus 1pt} 
{\normalfont} 
{0pt} 
{\itshape} 
{.} 
{.5em} 
{}

\newtheoremstyle{thmstylethree} 
{18pt plus 2pt minus 1pt} 
{18pt plus 2pt minus 1pt} 
{\normalfont} 
{0pt} 
{\bfseries} 
{.} 
{.5em} 
{}
\makeatother

\theoremstyle{thmstyleone}
\newtheorem{theorem}{Theorem}
\newtheorem{prop}[theorem]{Proposition} 

\theoremstyle{thmstylethree}

\theoremstyle{thmstyletwo}

\newtheorem{obs}{Observation}

\theoremstyle{thmstyleone}

\usepackage{xcolor} %
\usepackage[colorlinks=false,
            citebordercolor=green,
            pdfborder={0 1 2} 
           ]{hyperref}

\hypersetup{
    linkbordercolor=green, 
    urlbordercolor=green,
    filebordercolor=green,
}

\DeclareMathOperator*{\argmax}{argmax}

\makeatletter
\newlength\min@xx
\newcommand*\xxrightarrow[1]{\begingroup
  \settowidth\min@xx{$\m@th\scriptstyle#1$}
  \@xxrightarrow}
\newcommand*\@xxrightarrow[2][]{
  \sbox8{$\m@th\scriptstyle#1$}  
  \ifdim\wd8>\min@xx \min@xx=\wd8 \fi
  \sbox8{$\m@th\scriptstyle#2$} 
  \ifdim\wd8>\min@xx \min@xx=\wd8 \fi
  \xrightarrow[{\mathmakebox[\min@xx]{\scriptstyle#1}}]
    {\mathmakebox[\min@xx]{\scriptstyle#2}}
  \endgroup}
\makeatother

\title{Hyperoctant Search Clustering: A Method for Clustering Data in High-Dimensional Hyperspheres}

\author{
  Mauricio Toledo-Acosta \\
  Departamento de Matemáticas\\
  Universidad de Sonora (UNISON)\\
  Hermosillo, Son. 83000\\
  \texttt{mauricio.toledo@unison.mx} \\
    \And
  Luis Ángel Ramos García \\
  Departamento de Ciencias Computacionales e Inteligencia Artificial\\
  Universidad Autónoma del Estado de Morelos (UAEM)\\
  Cuernavaca, Mor. 62209 \\
  \texttt{luis.ramosgar@uaem.edu.mx} 
   \And
  Jorge Hermosillo Valadez\thanks{Corresponding author.} \\
  Departamento de Ciencias Computacionales e Inteligencia Artificial\\
  Universidad Autónoma del Estado de Morelos (UAEM)\\
  Cuernavaca, Mor. 62209 \\
  \texttt{jhermosillo@uaem.mx} \\
}

\begin{document}
\maketitle
\begin{abstract}
Clustering of high-dimensional data sets is a growing need in artificial intelligence, machine learning and pattern recognition. In this paper, we propose a new clustering method based on a combinatorial-topological approach applied to regions of space defined by signs of coordinates (hyperoctants). In high-dimensional spaces, this approach often reduces the size of the dataset while preserving sufficient topological features. According to a density criterion, the method builds clusters of data points based on the partitioning of a graph, whose vertices represent hyperoctants, and whose edges connect neighboring hyperoctants under the Levenshtein distance.

We call this method \emph{HyperOctant Search Clustering}. We prove some mathematical properties of the method. In order to as assess its performance, we choose the application of topic detection, which is an important task in text mining. 
Our results suggest that our method is more stable under variations of the main hyperparameter, and remarkably, it is not only a clustering method, but also a tool to explore the dataset from a topological perspective, as it directly provides information about the number of hyperoctants where there are data points. We also discuss the possible connections between our clustering method and other research fields.

\end{abstract}

\keywords{High-dimensional clustering \and Density \and Hyperoctants \and Graph Search \and Space Transformation.  }

\newpage
\section{Introduction}
\label{sec1}

Clustering is part of the exploratory data methods that aim at enhancing the practitioner’s natural ability to recognize patterns in the data being studied. Clustering allows to discover underlying structure, to gain insight into data, generate hypotheses, detect anomalies, and identify salient features. Also, it allows to perform natural classification, for instance, to identify the degree of similarity among forms or organisms (phylogenetic relationship); and to investigate compression methods for organizing the data and summarizing it through cluster prototypes. 

Cluster analysis of high-dimensional representations of unstructured data is a growing need in the context of artificial intelligence, machine learning and pattern recognition. In this paper, we propose a new clustering algorithm for points in high dimensional spaces endowed with the angular metric. This method is based on a combinatorial-topological approach applied to regions of space defined by signs of coordinates (hyperoctants), the method searches for clusters in a graph encoding the location of groups of points according to their coordinate signs. 

In the literature of Machine Learning, we can find many clustering approaches, which we can categorize essentially into two classes: partition-based methods, and hierarchical methods. Our method is an original combination of both. On the one hand, our method is density based, grouping hyperoctants together if they satisfy a condition regarding the density of the possible cluster. Therefore, our method is related to approaches such as DBSCAN \cite{Ester1996DBSCAN} and OPTICS \cite{ankerst_ACM99}. On the other hand, the proposed method builds and partitions a graph. Contrary to state-of-the-art hierarchical methods based on minimum spanning tree or dendograms \cite{jain88},\cite{Steinbach2004}, we construct a connected graph that follows the topology of the hyperoctants of the data space according to the Levenshtein distance. As Levenshtein distance is correlated to cosine similarity, we can work in non-euclidean and discrete spaces (e.g. binary or categorical feature spaces). We call our method HyperOctant Search Clustering (HOS-Clustering).

In high dimensional settings, points lying in a same hyperoctant are close to each other and two neighboring hyperoctants may have close points. We formally prove this assertion. In contrast to current approaches based on graphs, where nodes represent data items, and weighted edges represent similarities among the data items, each node in our graph represents a hyperoctant, connected to neighboring hyperoctants that change by only one dimension.  Hence, points lying on neighboring nodes in the graph, imply the closeness of these points in the feature space. 

Although classical hierarchical clustering allows for multiple granularity or to automatically discover nested structures \cite{Monath21Scalable}, our work allows to view data from a different perspective, as it allows for a topological exploration of data points based on overall scales, since the method allows to analyze the topology of the points from two global scales: (1) at the hyperoctant level, and (2) at the level of groups of hyperoctants. As the dimension grows, these global scales are adequate to account for the distribution of the data.

Thus, the proposed method extends the current possibilities of exploring high-dimensional data from a topological perspective. As the method is best suited for high-dimensional data sets containing positive and negative features, typical examples of which are dense vector representations produced by neural networks, the proposed method allows gaining insight into embedded representations of data.

Clustering has many useful applications, in this paper we selected a particular one in order to assess the performance of the proposed method. We tested our method in the task of topic detection. While there have been a significant number of innovative clustering methods in the field of image processing, whether for image segmentation \citep{shi2017novel}, or clustering purposes \citep{yellamraju2018clusterability}, novel text clustering methods beyond k-means \citep{sailaja2015overview, li2020text, mehta2021weclustering} are still missing. Over the last two decades, breakthrough neural network language models \citep{bengio2003neural, mikolov13, pennington2014glove, Vaswani2017, devlin-etal-2019-bert, zhangSemBERT2020} have enabled the derivation of embedded vector representations of text units of different kinds, which capture various syntactic and semantic properties of written language \citep{rogers-etal-2020-primer}. Interestingly, the purpose of the semantic spaces generated by these models bear a remarkable resemblance to theories on neural coding \citep{haxby2014decoding}. Moreover, these kind of embeddings have attracted the attention of other scientific disciplines, not only to investigate semantic phenomena related to language \citep{sassenhagen2020traces}, but also, for example, to represent genetic code \citep{du2019gene2vec}. Thus, there is indeed a huge potential in the discovery of new methods that allow to explore this type of computational representations more effectively. Given that our method is a density-based clustering method, we compare its results to those produced by DBSCAN as baseline. The comparison was done in three different text corpora.

The rest of this paper is organized as follows: In Section \ref{sec:rw},  we situate our contribution with respect to related work. In Section \ref{sec:methods}, we formally describe the clustering method, and the experimental design to assess its performance. In Section \ref{sec:experiments}, we present the experimental results, that we discuss in Section \ref{sec:discussion}. We conclude the paper in Section \ref{sec:conc} where we summarize our contributions.

\section{Related work}
\label{sec:rw}

Clustering methods can be categorized into three classes: Hierarchical, Partitioning and Density based. Hierarchical techniques produce a nested sequence of partitions, with a single, all-inclusive cluster at the top and singleton clusters of individual points at the bottom. Depending on whether an ascending or descending process is followed, the clustering is either ``agglomerative" or ``divisive" \citep{kohn2014hierarchical}. Partitioning algorithms can be better understood from the notion of distance and neighbor. 
The best known algorithms in this class are \emph{K-means} \citep{macqueen1967}, with its variants such as Spherical K-Means \citep{dhillon2001concept}, Multi-Cluster Spherical K-Means \citep{tunali2016improved} or Robust and Sparse K-Means \citep{brodinova2019robust}, and  \emph{K-nearest-neighbors} \citep{fix1989discriminatory}. Both approaches have been widely reported in the literature, with an emphasis on improving computational performance and reducing complexity
 \citep{ chen2009FastKNN, zhang2012graph, li2019approximate}. Density based clustering aims at finding clusters at different levels of granularity with appropriate noise filtering, depending on the notion of density of a region \citep{ertoz2003finding}. Over the last two decades, numerous density based clustering techniques have been proposed (for example, see \cite{bhattacharjee2021survey} for a recent review).

For clustering purposes, the most relevant aspect of the curse of dimensionality concerns its effect on distance or similarity: a finite set of data points becomes increasingly ``sparse” as the dimensionality increases, thus distances between points become relatively uniform \citep{beyer1999}. Hence, meaningful clusters are harder to detect as distances are increasingly similar for growing dimensionality.

One approach to deal with this phenomenon is to reduce the dimensionality of the data. Some techniques for clustering high dimensional data have included both feature transformation and feature selection techniques. Feature transformation algorithms employ dimensionality reduction to project data into a lower dimensional meaningful space, thus uncovering latent structure in data \citep{bouveyron2007high}. In this category, we can include the learning to hash or binary code learning methods \citep{gong2015web, wang2017survey, shen2017asymmetric, zhang2018binary}. Recently, binary encoding has been combined with classical approaches to clustering, such as k-means \citep{shen2017compressed}, or k-nearest-neighbors \citep{fan2020efficient}. Feature selection methods select only the most relevant of the dimensions to reveal groups of objects that are similar on only a subset of their attributes. Subspace clustering \citep{agrawal1998automatic, cheng1999entropy, yan2020efficient, chen_coupled_2022} is an extension of feature selection that attempts to find clusters in different subspaces of the same data set \citep{kriegel2009}. 

When working with embedded representations, searching for clusters in the full feature set is important, as it has been empirically shown that these representations cluster by semantic similarity (topic similarity), which is a property captured by the full vector. Hence, the search for clusters in the whole set of features is key, and thus feature selection or feature transformation methods are often not adequate for clustering text entities. In this case, density-based methods such as DBSCAN are more effective. However, hyper-parameter tuning is not a trivial task when dealing with high-dimensional data, and therefore, the correct use of these methods is challenging \citep{schubert2017DBSCAN,xu2005survey}.

It has been recognized that the similarity measure is a key factor for clustering. However, it is still challenging for existing similarity measures to cluster non-spherical data with high noise levels \cite{huang_adaptive_2020}. From a theoretical point of view, the way we use Levenshtein distance is isomorphic to the Hamming distance, as our $\{+,-\}^D$ labels are isomorphic to $\{0,1\}^D$ strings. The relevance of Hamming distance in the context of pattern recognition has been subject of theoretical analysis in the past \cite{gasieniec_approximation_2004}. Furthermore, it has been shown that for very small distances, it is possible to demonstrate the equivalence of distance measures, commonly employed in the context of clustering (e.g. hamming distance), in order to compare partitions created by different methods \cite{meila_local_2012}. 


Furthermore, it has been shown that most clustering tasks are NP-hard \citep{ackerman09a}, and that for a set of three simple properties: scale invariance; richness, the requirement that all partitions are realisable; and a consistency condition on the reduction and stretching of individual distances, there is no clustering function that satisfies all three \citep{kleinberg02}. This indicates that basic trade-offs are inherent to any solution of a clustering problem. 

In these regards, we provide evidence that our method is a sound alternative to perform clustering tasks in high-dimensional spaces endowed with the angular metric, with the additional advantages that it allows to acquire insight about some topological properties of the semantic space, and also, its performance is more stable under variations of its most important hyperparameter.

\section{Methods}
\label{sec:methods}

HOS-Clustering is a density-based clustering algorithm. The method regards clusters as high density regions separated by neighborhoods of low density. The search for these regions is performed in terms of the location of the hyperoctants containing points of the data set; that is, we consider the signs of the coordinates of each point. As a consequence of this loosely constrained approach, clusters found by HOS-Clustering do not have special geometric restrictions, such as convexity in other clustering methods. Our method assumes that the underlying metric in the data set space is the angular metric, or equivalently, the cosine similarity. 

We start with the necessary definitions and background. Then we detail our clustering method together with the tools we use to assess its performance in topic detection tasks.

\subsection{Notation and Background}
\label{subsec:notation}

Consider the metric space $\left(\mathbb{R}^D\setminus\{0\},d\right)$, where $d:\left[-1,1 \right]\rightarrow [0,\pi]$ is the angular metric given by 

\begin{equation*}
d(u,v)=\arccos\left( \frac{u\cdot v}{\lvert u \lvert\cdot \lvert v\lvert} \right) \;.    
\end{equation*}

\noindent Here, $u\cdot v$ denotes the dot product in $\mathbb{R}^D$, and $|z|$ denotes the Euclidean norm of $z\in \mathbb{R}^D$. Since $d(r_1u,r_2v)=d(u,v)$ for any $r_1,r_2>0$, we only consider points lying in the unit sphere $\mathbb{S}^{D-1}\subset \mathbb{R}$. Recall the cosine similarity given by

\begin{equation*}
\text{sim}(u,v) = \frac{u\cdot v}{\lvert u \lvert\cdot \lvert v\lvert}\;.
\end{equation*}

\noindent The \emph{diameter} of a finite set $B\subset\mathbb{R}^D$ is defined as 

\begin{equation*}
\text{diam}(B)=\max \{ d(x,y)\; \lvert \;x,y\in B\}\;.
\end{equation*}

\noindent Let the \emph{linear density} (or density) of a finite set $B\subset\mathbb{R}^D$, $|B|>1$, be:
    \begin{equation}
    \label{eq:linear-density}
    \delta(B)=\frac{\lvert B \lvert}{\text{diam}(B)}\;,
    \end{equation}
    
\noindent where $|B|$ denotes the number of elements of $B$. This function $\delta$ accounts for the density of the set $B$, in terms of how many elements it contains with respect to the maximum distance between the elements. Typically, the density is measured in terms of elements per $D$-dimensional units of volume. However, in order to avoid dealing with the term $\text{diam}(B)^D$ which can be very small, or very large, as the dimensionality $D$ increases, we consider instead the linear density given by Equation (\ref{eq:linear-density}).

\subsubsection{Hyperoctants}
\paragraph{Hyperoctants as strings of signs.}
We consider the space $\mathbb{R}^D\setminus\{0\}$ divided into $2^D$ pairwise disjoint open regions, called hyperoctants. A hyperoctant is a region of space defined by one of the $2^D$ possible combinations of coordinate signs $(\pm, ... ,\pm)$. For example, in $\mathbb{R}^3$, the combinations $+++$ and $-+-$ determine the following hyperoctants 

\begin{align}
+++:\;\; & \{(x_1,x_2,x_3)\in \mathbb{R}^3 \, \mid \, x_1,x_2,x_3 >0 \}  \nonumber \\  
-+-:\;\; & \{(x_1,x_2,x_3)\in \mathbb{R}^3 \, \mid \, x_1,x_3 <0,\,x_2>0 \} \label{eq:example_2hos}
\end{align}

Observe that points with some coordinate equal to 0 are not contained in any hyperoctant according to this definition. However, we do not expect this to be a problem since 0-valued coordinates in the embeddings produced by neural networks are very unusual. 

In low dimensional cases there are only a few hyperoctants (for example, there are only 4 quadrants in $\mathbb{R}^2$). However, as the dimension $D$ increases, the number of hyperoctants increases exponentially. In this case, the size of any finite set of points $|A|$ becomes much smaller than the number of hyperoctants. This means that the string of signs defining the hyperoctant of each point is a useful label for its location in space. At the same time, as $D$ increases, whenever two points are in the same hyperoctant, it means that they are close to each other, and probably have to be in the same cluster. This observation is formalized as follows.

\begin{obs}\label{obs:points_same_HO}
Let $S_{D}$ be the surface area of $\mathbb{S}^{D-1}$. We denote by $S^{\text{HO}}_{D}$ the surface area of the intersection of each hyperoctant and $\mathbb{S}^{D-1}$. Since there are $2^{D}$ hyperoctants, $S^{\text{HO}}_{D}$ is given by 

$$
S^{\text{HO}}_{D}=\frac{S_D}{2^{D}}= 
\begin{cases}
\frac{\pi^k (k-1)!}{(2k-1)!}, & \text{ if } D=2k\text{, for some integer }k \\
\frac{\pi^{k+1}}{4^k k!}, & \text{ if } D=2k+1\text{, for some integer }k. 
\end{cases}
$$

It is straightforward to verify that $\lim_{D\rightarrow\infty}S^{\text{HO}}_{D} = 0$. This means that, as the dimension $D$ increases, the region of space determined by each hyperoctant gets smaller with respect to the total area of $\mathbb{S}^{D-1}$. Therefore, as $D$ increases, the fact that two points lie in the same hyperoctant means that they are close to each other.  
\end{obs}

\paragraph{Distance between hyperoctants.}
We will denote each hyperoctant by its respective ``\emph{sign label}'' as we did in the example described in (\ref{eq:example_2hos}). In this regard, a sign label is an element of $\{+,-\}^D$; that is, a string of size $D$ over the alphabet $\{+,-\}$. To determine the proximity of two hyperoctants, in a way that reflects the angular distance between the points located at these hyperoctants, we use the Levenshtein distance $d_L$ between their sign labels  \citep{levenshtein1966binary}. This distance is defined as the number of places where the two strings are different. For example, in $\mathbb{R}^3$,
    $$d_L\left(+++,-+- \right) = 2.$$

In the sequel, we will refer to a sign label by the symbol $\phi$. It is worth noting that the sign label $\phi$ defining the hyperoctant containing some point $v\in\mathbb{R}^D$ is directly related to the sign vector of $v=(v^{(1)},...,v^{(D)})$, defined as $\sigma_v = (\sigma_v^{(1)},...,\sigma_v^{(D)})$ where

$$\sigma_v^{(k)} = 
\begin{cases}
1, & v^{(k)} \geq 0 \\
-1, & v^{(k)} < 0\;.
\end{cases}
$$

\noindent Alternatively, the sign vector is given by 

$$
\sigma_v = \left(\frac{v^{(1)}}{\lvert v^{(1)}\lvert},...,\frac{v^{(D)}}{\lvert v^{(D)}\lvert} \right)\;.
$$

We will denote the one-to-one correspondence between a sign vector and its sign label by writing: $\sigma \sim \phi$. Thus, we use indistinctly the sign vector, or sign label, to compute Levenshtein distances between hyperoctants, and sign labels for identifying each hyperoctant during the clustering process.

For any point $v\in\mathbb{R}^D$, we denote by $h:\mathbb{R}^D\rightarrow \{+,-\}^D$ the function mapping $v$ to the the sign label defining the hyperoctant containing $v$. Thus we have:

\begin{figure}[htbp]
    \centering
    \includegraphics[scale=0.15, ]{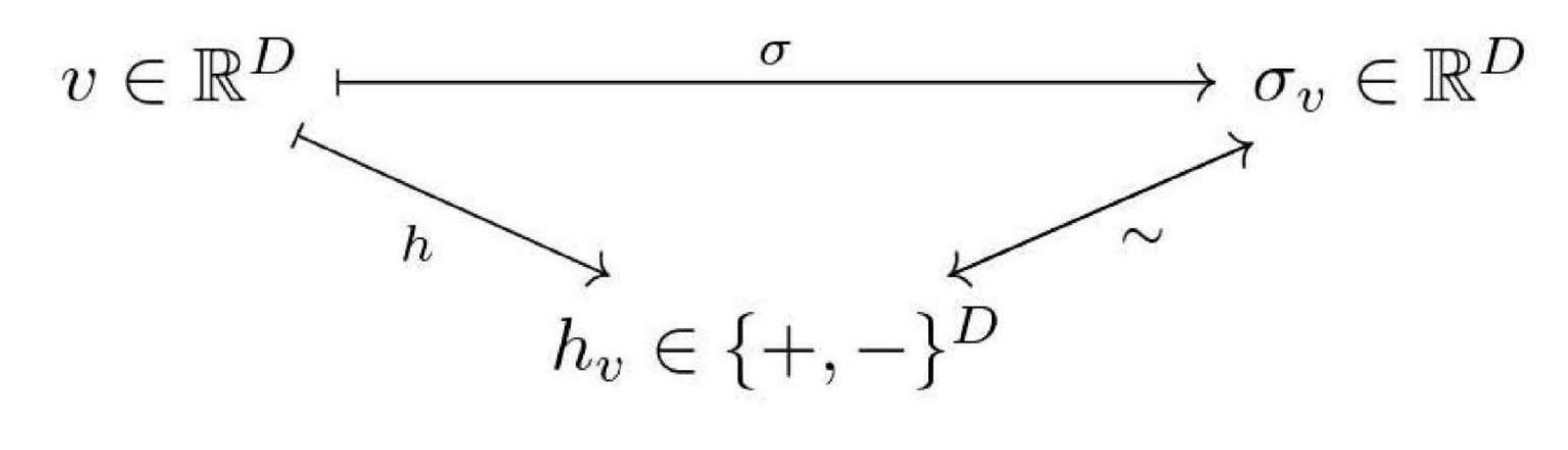}
\end{figure}

For example, if $v=(-3,2)$, then $\sigma_v=(-1,1)$, and the sign label is given by $h_v=-+$. 

If we consider two points $v,w\in\mathbb{R}^D$, we can define the Levenshtein distance between them by
$$
d_L(u,v):=d_L(h_u,h_v)\;.
$$

Now, we describe a way to compute all Levenshtein distances between all pairs of row vectors in a matrix. Let $A$ be a matrix of $N$ row vectors $A\in\mathcal{M}_{N\times D}\left(\mathbb{R}\right)$ given by

\begin{equation*}
A=\left(\begin{array}{c}
     v_1  \\
     ... \\
     v_N \\
\end{array}\right).
\end{equation*}
    
The Levenshtein distance $d_L$ between two row vectors $v=(v^{(1)},...,v^{(D)})$ and $w=(w^{(1)},...,w^{(D)})$ can be computed as follows

\begin{equation}\label{eq:defn_lev}
  d_L(v,w)= \frac{D-\sum_{i=1}^{D}\frac{v^{(i)}}{\lvert v^{(i)}\lvert}\frac{w^{(1)}}{\lvert w^{(i)}\lvert}}{2}\;.   
\end{equation}

We denote by $A_\sigma$ the matrix consisting of the sign row vectors. Using Equation (\ref{eq:defn_lev}), we can define a matrix $\Delta_L$, whose entry $i,j$ is the Levenshtein distance $d_L(v_i,v_j)$, as follows

\begin{equation*}
    \Delta_L = \frac{1}{2}\left( \mathbf{D} - A_\sigma \cdot A_\sigma^T \right)\;,
\end{equation*}
\noindent where $\mathbf{D}\in \mathcal{M}_{N\times N}(\mathbb{N})$ is the matrix with every entry equal to $D$.

\begin{figure}[htbp]  
\centering 
  \begin{subfigure}[t]{0.46\linewidth}
    \includegraphics[width=0.8\linewidth, ]{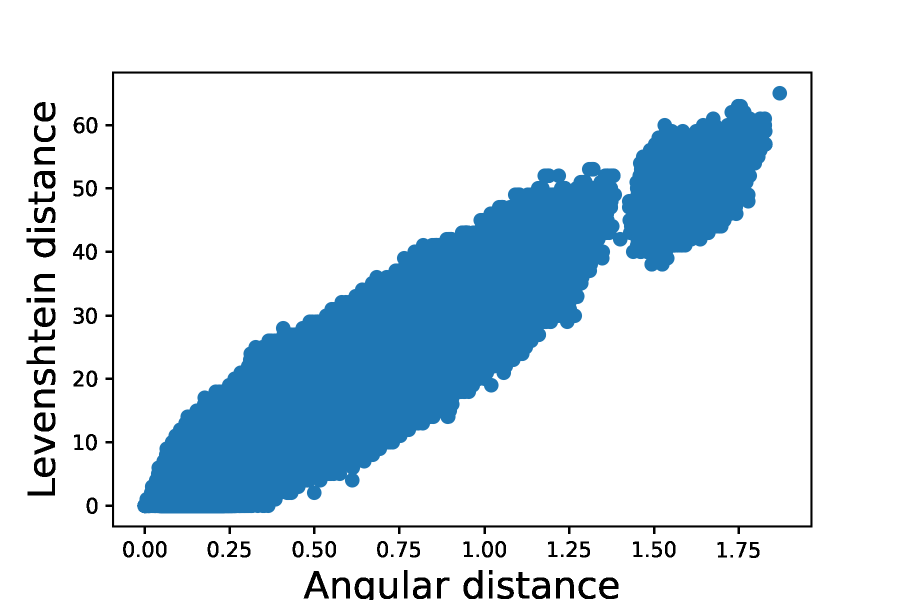}
    \caption{Angular and Levenshtein distances with \texttt{word2vec} embeddings.}
    \label{fig:correlations_1_anglev}
  \end{subfigure}
  \begin{subfigure}[t]{0.46\linewidth}
    \includegraphics[width=0.8\linewidth, ]{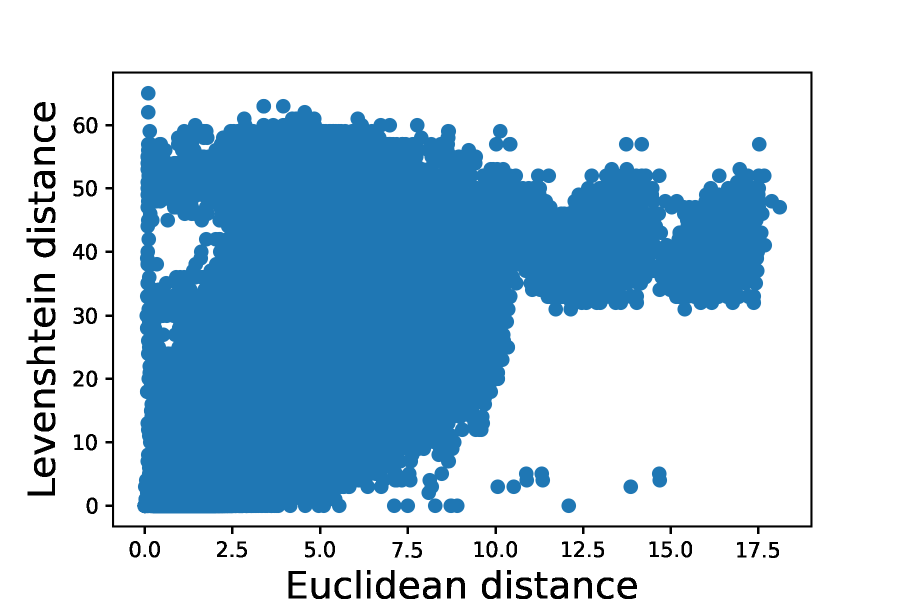}
    \caption{Euclidean and Levenshtein distances with \texttt{word2vec} embeddings.}
    \label{fig:correlations_1_euclev}
\end{subfigure}

  \begin{subfigure}[t]{0.46\linewidth}
    \includegraphics[width=0.8\linewidth, ]{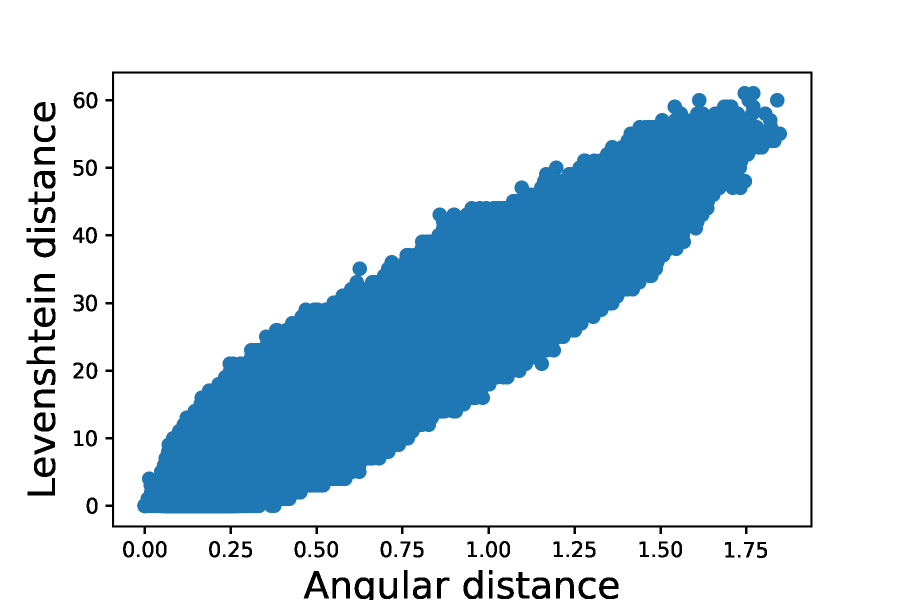}
    \caption{Angular and Levenshtein distances with \texttt{FastText} embeddings.}
    \label{fig:correlations_2_anglev}
  \end{subfigure}
  \begin{subfigure}[t]{0.46\linewidth}
    \includegraphics[width=0.8\linewidth, ]{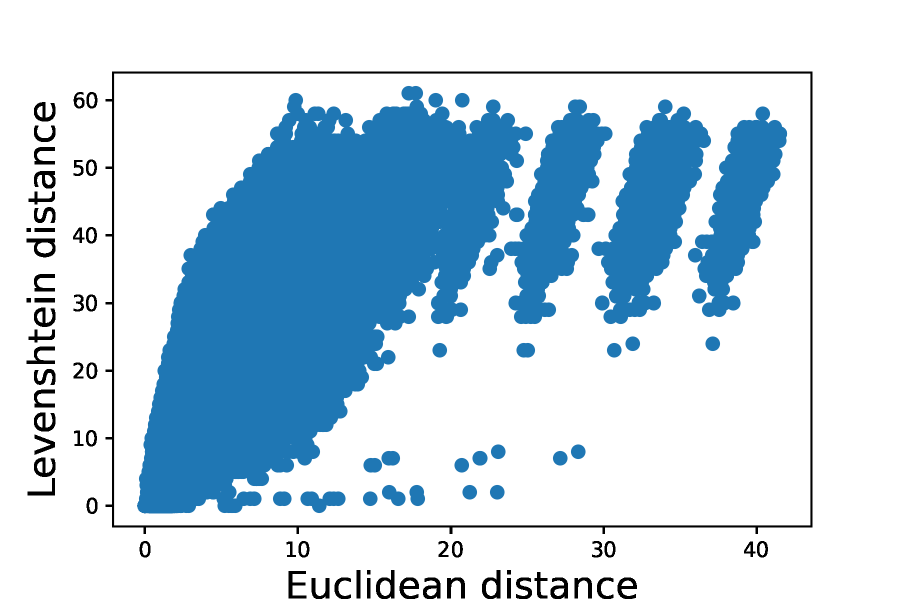}
    \caption{Euclidean and Levenshtein distances with \texttt{FastText} embeddings.}
    \label{fig:correlations_2_euclev}
\end{subfigure}

\caption{The Levenshtein distance can be a good substitute for the angular distance (left), but not so much for the Euclidean distance (right). In both cases data points correspond to either \texttt{word2vec} \citep{mikolov2013efficient} or \texttt{FastText} \citep{bojanowski2016enriching} embeddings.}
\label{fig:correlations}
\end{figure}

Evidence suggests that there is a correlation between the angular distance between pairs of points, and the Levenshtein distance between their respective hyperoctant sign labels (see Figures \ref{fig:correlations_1_anglev} and \ref{fig:correlations_2_anglev}). This is also supported by the fact that $d_L(\sigma_u,\sigma_v)+  u\cdot v\leq D$. Because of this, one can replace the notion of \emph{closeness} using the angular distance between points with the Levenshtein distance between their respective sign labels. In Section \ref{subsubsec:centering_value} we will analyze the details of this substitution.

On the other hand, there is a weaker correlation between the Euclidean distance and the Levenshtein distance (see Figures \ref{fig:correlations_1_euclev} and \ref{fig:correlations_2_euclev}). All four sub-figures in Figure \ref{fig:correlations} correspond to two particular data sets of word embeddings, which have been included here without further details for illustrative purposes.

\begin{figure}[htbp]
    \centering
    \includegraphics[width=0.3\linewidth, ]{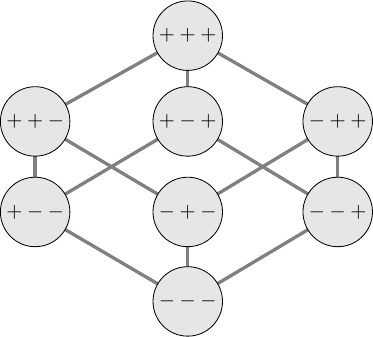}
    \caption{The graph $P_D$ for $D=3$ representing the spatial location of the 8 octants of $\mathbb{R}^3$.}
\label{fig:grafo_PD}
\end{figure}

\paragraph{Encoding hyperoctants' proximity using graphs.} The set of hyperoctants of a $D$-dimensional space has a combinatorial structure which encodes the location of each hyperoctant with respect to the others. We can encode this information in the following graph $P_D=(V,E)$. The set $V$ is the set of hyperoctants labeled by their respective sign label $\phi$. Two sign labels $\phi_1$ and $\phi_2$ are connected by an edge in $E$ if the Levenshtein distance between $\phi_1$ and $\phi_2$ is 1. In Figure \ref{fig:grafo_PD} we show an example of the graph $P_D$ for $D=3$.

As the number of nodes and edges of $P_D$ increases exponentially with $D$, we restrict our attention to a \emph{smaller} graph. Let $A\subset\mathbb{R}^D$ be a finite set, we will define a weighted graph $G_0=(V_0,E_0)$ as follows:

\begin{itemize}
    \item The set $V_0$ is the set of sign labels, defining the hyperoctants containing points of $A$. That is
        $$V_0 = \{ h_v \; \mid \; v\in A\}\;.$$
    \item Two sign labels $\phi_1$ and $\phi_2$ are connected by an edge in $E_0$ if, either $\overline{\phi_1\phi_2}\in E$ or the following condition holds: any minimal length path between $\phi_1$ and $\phi_2$ in $P_D$ contains no elements of $V_0$. The weight associated to the edge is $d_L(\phi_1,\phi_2)$.
\end{itemize}

\begin{figure}[htbp]

\centering
\begin{subfigure}[t]{.25\linewidth}
    \scalebox{0.55}{
    \includegraphics{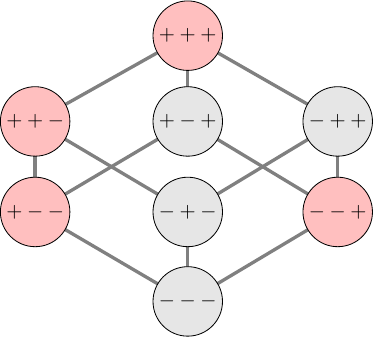}}
    \caption{The graph $P_D$ for $D=3$.}
    \label{fig:reduced-graph-examples-1}
    \end{subfigure}
    \hspace{4em}
    \begin{subfigure}[t]{.25\textwidth}
    \scalebox{0.6}{
    \includegraphics{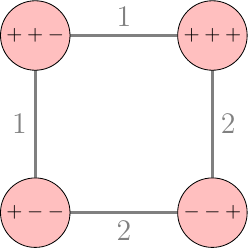}
    }
    \caption{The graph $G_0$.}
    \label{fig:reduced-graph-examples-2}
    \end{subfigure}
    \begin{subfigure}[t]{.25\textwidth}
    \scalebox{0.6}{
    \includegraphics{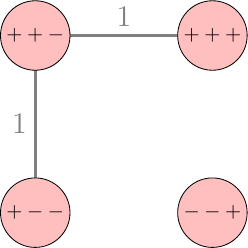}
    }
    \caption{The graph $G_0^{d_0}$ for $d_0=1$.}
    \label{fig:reduced-graph-examples-3}
    \end{subfigure}
    \caption{Construction of the graphs $G_0$ and $G_0^{d_0}$, the graph obtained by deleting the edges of $G_0$ such that $d_L(\phi_1,\phi_2)>d_0$. Elements of $V_0$ (that is, hyperoctants containing data points) are colored in red.}
    \label{fig:reduced-graph-examples}
\end{figure}

We call the graph $G_0$, the \emph{reduced graph} of $A$. In Figure \ref{fig:reduced-graph-examples}, we show an example of the construction of this graph. We start with the 8 octants of $\mathbb{R}^3$, given by their respective signs, represented in the graph $P_D$. The set $V_0=\{+--,++-,--+,+++\}$ is depicted in red, both as a set of nodes of $P_D$ (Figure \ref{fig:reduced-graph-examples-1}), and as a set of nodes of $G_0$ (Figure \ref{fig:reduced-graph-examples-2}).

We have the following propositions, the first of which is straight-forward to verify.

\begin{prop}
Any minimal length path in $P_D$ between $v$ and $w$ has length $d_L(v,w)$.
\end{prop}

\begin{prop}
The graph $G_0$ is connected.
\end{prop}

\begin{proof}
Let $\phi,\psi\in V_0$ be two sign labels. If $\overline{\phi\psi}\in E$, then $\overline{\phi\psi}\in E_0$. If $\overline{\phi\psi}\not\in E$, we have two cases:

\begin{itemize}
    \item Any minimal length path between $\phi$ and $\psi$ in $P_D$ contains no elements of $V_0$. It follows that $\overline{\phi\psi}\in E_0$.
    \item There exists a minimal length path between $\phi$ and $\psi$ in $P_D$ containing an element $y\in V_0$. In this case, let $\mathcal{P}$ be the collection of such minimal length paths. For the sake of clarity, we will consider that any path $q \in \mathcal{P}$ is a collection of nodes that we denote 
        $$q=\{\phi=y_0,y_1,...,y_{k-1},y_k=\psi\}\;.$$
    Let $q\in \mathcal{P}$ be such that $\lvert V_0\cap q \lvert \geq \lvert V_0\cap p \lvert$ for any $p\in\mathcal{P}$. We now show that $\overline{y_0 y_1},...,\overline{y_{k-1} y_k}\in E_0$. Let us assume that $y_j$ and $y_{j+1}$ are not connected by an edge in $E_0$, then there exists a minimal length path in $P_D$ between $y_j$ and $y_{j+1}$ containing an element $x\in V_0$. Therefore, the path
    $$q'=\{\phi=y_0,y_1,...,y_j,x,y_{j+1},...,y_{k-1},y_k=\psi\}$$
    has length $d_L(\phi,\psi)$, and $q\subset q'$ contradicting that $q$ is maximal. Therefore, 
    $$\overline{y_0 y_1},...,\overline{y_{k-1} y_k}\in E_0$$ 
    and thus, there is a path in $G_0$ between $\phi$ and $\psi$. This shows that $G_0$ is connected.
\end{itemize}
\end{proof}

This proposition means that the graph $G_0$ on which we will perform our search is always connected and, therefore, all points are \emph{reachable} by some walk. This situation is desirable if we want to consider all points in a data set in the clustering process.

However, if the data set contains highly isolated points, and given that the graph $G_0$ is connected, these isolated points will be connected to their nearest neighbours, no matter how far they are. This, in turn, might introduce noise to the clustering algorithm. This behaviour can be compensated by imposing a maximum distance threshold $d_0$ on the graph $G_0$, thus disconnecting the points that are too far apart and discarding them from the clustering process. We denote by $G_0^{d_0}$ the graph $G_0$ obtained by deleting the edges of $G_0$ whose weight is greater than $d_0$. By appropriately choosing $d_0$, the hyperoctants containing points far apart are disjointed nodes in this new graph $G_0^{d_0}$. 

Observe that $G_0 = G_0^{D}$. Therefore, we will use the graph $G_0^{d_0}$, with $d_0$ being a hyperparameter depending on the nature of the data set. Typically, we choose $d_0$ greater than the average edge weight in $G_0$. Figure \ref{fig:reduced-graph-examples-3} shows an example of $G_0^{d_0}$. 

\subsubsection{Centering Value for the data set}
\label{subsubsec:centering_value}

\begin{figure}[htbp]
    \centering
    \begin{subfigure}[t]{.3\linewidth}
    \includegraphics{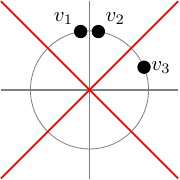}
    \caption{Non-centered data set.}\label{fig:non-centered}
    \end{subfigure}
    \hspace{4em}
    \begin{subfigure}[t]{.3\linewidth}
    \includegraphics{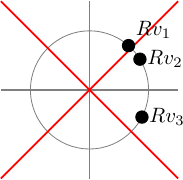}
    \caption{Centered data set.}\label{fig:centered}
    \end{subfigure}
    
    \caption{The same set of points under different coordinates.}
    \label{fig:centered-non-centered-data set}
\end{figure}

Now, we turn our attention to the question of whether the sign label of each hyperoctant adequately represents the points in the hyperoctant. In order to increase the effectiveness of the algorithm by replacing the notion of proximity between points, using the angular metric, by the notion of the Levenshtein distance between hyperoctant labels, it is desirable that as many points in the data set as possible are \emph{centred} within their respective hyperoctants. To illustrate this, note Figure \ref{fig:non-centered}, where it can be seen that points $v_1$ and $v_2$ are more likely to belong to the same cluster, but the points are not \emph{optimally} centered on their respective hyperoctants. After applying an adequate rotation $R$ around the origin, Figure \ref{fig:centered} shows that the same points are now located closer to the center of their respective hyperoctants.

This notion can be quantified by measuring, via the cosine similarity, how close each point $v$ is to the middle point of its respective hyperoctant. Given an hyperoctant defined by its sign label $\phi$, we say that its middle point is a point $\sigma_v$ such that $\sigma_v\sim\phi$.

$$\text{sim}(v,\sigma_v) = \frac{1}{\sqrt{D}}\sum_{j=1}^D \lvert v^{(j)}\lvert\;.$$

Given a finite set $A=\{v_1,...,v_{N}\}\subset \mathbb{S}^{D-1} \subset \mathbb{R}^D$, with $v_i=\left(v_i^{(1)},...,v_i^{(D)} \right)$, the average similarity between elements of $A$ and the respective middle point of their hyperoctant is given by

\begin{equation*}
    \mathbf{c}(A)= \frac{1}{N}\sum_{k=1}^N \text{sim}(v_k,\sigma_{v_k})=
    \frac{1}{N\sqrt{D}}\sum_{k=1}^N \sum_{j=1}^D \lvert v_k^{(j)}\lvert\;.
\end{equation*}

We call $\mathbf{c}(A)$ the \emph{centering value} of a finite set $A\subset \mathbb{S}^{D-1}$, and we use it as a measure quantifying the average centering of the elements of $A$. In Figure \ref{fig:centered-non-centered-data set}, the set of points in \ref{fig:non-centered} has a lower centering value than the set of points in \ref{fig:centered}.

\subsubsection{Rotations in \texorpdfstring{$\mathbb{R}^D$}{RD}} 

A rotation is a distance-preserving transformation of a $D$-dimensional Euclidean space. Equivalently, a rotation can be regarded as a $D\times D$ orthogonal matrix. Observe that a rotation is also a symmetry of the unit sphere $\mathbb{S}^{D-1}$ endowed with the angular metric. Up to a change of basis, a rotation is given by a matrix

$$R_\Theta = \left(
\begin{array}{cccccc}
R_{\theta_1} & & & & & \\
& \ddots & & & \mathbf{0} & \\
& & R_{\theta_k} & & & \\
& & & 1 & & \\
& \mathbf{0} & & & \ddots & \\
& & & & & 1
\end{array}
\right),$$
\noindent where $\Theta=\{\theta_1,...,\theta_k\}$ is a collection of $k$ angles $\theta\in[0,2\pi)$, and $R_\theta$ is a 2-dimensional rotation matrix with angle $\theta$. Observe that $k\leq \lfloor \frac{D}{2} \rfloor$. In general, a rotation matrix is given by
$$
R_{\Pi,\Theta} = P R_\Theta P\;,
$$
\noindent where $\Pi$ is a set of $k$ pairs $(i,j)$, each of these indicating which coordinates are rotated with the respective angle $\theta$. The matrices $P$ are permutation matrices determining the change of coordinates. 

For example, the 5-dimensional rotation matrix $R_{\Pi,\Theta}$ with $\Pi=\{(1,3),(2,5)\}$ and $\Theta = \{\frac{\pi}{4},\frac{\pi}{6}\}$ is explicitly given by

$$R_{\Pi,\Theta} = \left(
\begin{array}{ccccc}
\cos \frac{\pi}{4} & 0 & -\sin\frac{\pi}{4} & 0 & 0 \\
0 & \cos \frac{\pi}{6} & 0 & 0 & -\sin \frac{\pi}{6}\\
\sin\frac{\pi}{4} & 0 & \cos \frac{\pi}{4} & 0 & 0\\
0 & 0 & 0 & 1 & 0\\
0 & \sin \frac{\pi}{6} & 0 & 0 & \cos \frac{\pi}{6}
\end{array}
\right).$$

Given a finite set of points $A$, we seek a rotation matrix $R_{\Pi,\Theta}$ maximizing the value $\mathbf{c}(R_{\Pi,\Theta} A^T)$. This rotation matrix is parameterized by a pair $(\Pi,\Theta)$ given as before. Once we find the adequate rotation matrix, we can replace the set $A$ by $R_{\Pi,\Theta} A^T$ to obtain a better centered set of points in the sense of Figure \ref{fig:centered-non-centered-data set}.

\subsection{The Clustering Method}
\label{subsec:the-method}

Now, we describe the proposed clustering method using the notation and ideas developed in Subsection \ref{subsec:notation}. 

We start with a $D$-dimensional data set of size $N'$, which is a finite set $$A=\{v_1,...,v_{N'}\}\subset \mathbb{S}^{D-1} \subset \mathbb{R}^D\;.$$ 

Using an optimization method (for example, local search of global optimum like simulated annealing), we first find a rotation $R_{P,\Theta}$ maximizing the centering value $\mathbf{c}(A)$ of the data set. We replace the set $A$ by $R_{\Pi,\Theta} A^T$, and denote this new set again by $A$. 

We identify each point $v\in A$ by its sign label $h_v\in\{+,-\}^D$, which defines an hyperoctant in $\mathbb{R}^D$. From the construction of the reduced graph $G_0$, recall the set of sign labels
\begin{equation*}
     V_0 = \{ h_v \; \mid \; v\in A\}\;.
\end{equation*}

Since we can have two or more points in a same hyperoctant, it follows that
\begin{equation}
    \label{eq:ho_size_reduction}
    N:=\lvert V_0\lvert \leq \lvert A\lvert=N'\;.
\end{equation}
\noindent It is worth observing that, in practice, the rotation of the data set increments the difference $N'-N$, thus reducing the size of the data set, which improves the performance of the clustering method. 

From this point on, two points lying in the same hyperoctant will be considered as the same instance during the clustering process. In other words, at the end of the process, if one point is in a cluster, all other points in the same hyperoctant will be in the same cluster. The motivation behind this principle was given by Observation \ref{obs:points_same_HO}.

\begin{figure}[htbp]
    \centering
    \begin{subfigure}[t]{.3\linewidth}
    \scalebox{1}{
    \includegraphics[width=.7\linewidth, ]{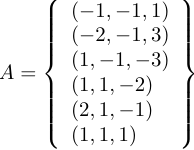}
    }
    \caption{The data set}
    \end{subfigure}
    \hspace{2em}
    \begin{subfigure}[t]{.5\linewidth}
    \centering
    \scalebox{0.65}{
    \includegraphics[width=\linewidth, ]{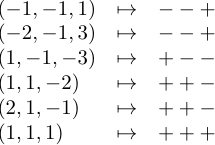}}
    \caption{Data points and corresponding sign labels.}
    \end{subfigure}
    \vspace{2em}
    
    \begin{subfigure}[t]{.3\linewidth}
    \centering
    \scalebox{0.82}{
    \includegraphics[width=\linewidth, ]{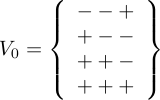}}
    \caption{The set of sign labels $V_0$}
    \end{subfigure}
    \hspace{4em}
    \begin{subfigure}[t]{.5\linewidth}
    \centering
    \scalebox{0.55}{
    \includegraphics[width=\linewidth, ]{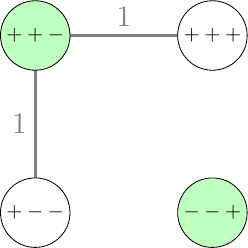}}
    \caption{The graph $G_0^{1}$. The set of proto-clusters $W_0$ is depicted in green.}
    \end{subfigure}
    \vspace{2em}
    
    \begin{subfigure}[htbp]{.3\linewidth}
    \centering
    \scalebox{1}{
    \includegraphics[width=.7\linewidth, ]{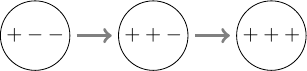}}
    \caption{The walk obtained via BFS, assuming $\varphi_0=+--$.}
    \end{subfigure}
    \hspace{4em}
    \begin{subfigure}[t]{.5\linewidth}
    \centering
    \scalebox{0.75}{
    \includegraphics[width=\linewidth, ]{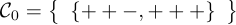}}
    \caption{Suppose that the walk produces the collection $\mathcal{C}_0$.}
    \end{subfigure}
    \vspace{2em}
    
    \begin{subfigure}[t]{0.3\linewidth}
    \centering
    \scalebox{1.25}{
    \includegraphics[width=.7\linewidth, ]{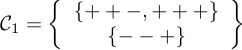}}
    \caption{We form $\mathcal{C}_1$ by adding to $\mathcal{C}_0$ elements of $W_0$ not present in $\mathcal{C}_0$.}
    \end{subfigure}
    \hspace{4em}
    \begin{subfigure}[t]{0.5\linewidth}
    \centering
    \includegraphics[width=.7\linewidth, ]{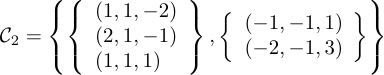}
    \caption{The collection of clusters $\mathcal{C}_2$, taking $k_0=2$. }
    \end{subfigure}
    
    \caption{Walk-through example illustrating the clustering process for a data set $A$ with no rotation performed.}
    \label{fig:prin}
\end{figure}

The clustering principle starts by identifying hyperoctants containing more than one point. These subsets of points are \emph{potential} clusters (we call them `proto-clusters'). Then, we construct $G_0^{d_0}$ and perform a walk through this graph starting from the node with minimal combined Levenshtein distance to every other node. As we walk, we group nodes together according to a density criterion. The result is a collection of subsets of nodes of $G_0^{d_0}$, possibly including some proto-clusters. These subsets together with all proto-clusters not included in the previous collection, will become true clusters if they verify a cardinality condition. Figure \ref{fig:prin} depicts the process for a simple data set, where some assumptions are made for illustration purposes.

We now describe the formal details of the process. We denote by $\varphi_0\in V_0$ the sign label with the minimal combined Levenshtein distance to every other element of $V_0$. That is,

\begin{equation*}
    \varphi_0 = \underset{\phi\in V_0}{\text{argmin}}  \sum_{\psi\in V_0} d_L(\phi,\psi)\;.
\end{equation*}

We denote by $P_\phi$ the set of points in $A$ lying in the hyperoctant defined by the sign label $\phi$. Formally,  
\begin{equation*}
P_{\phi} = \{ v \in A \; \big\lvert \; \sigma_v \sim \phi \}\;.      
\end{equation*}

As previously stated, the step shown in inequality (\ref{eq:ho_size_reduction}), induces a reduction of the number of entities. At the same time, it forms the first \emph{proto-clusters}, which later might be enriched with new elements. By a \emph{proto-cluster} we mean a set $P_\phi$ such that $\lvert P_\phi \lvert>1$, for some $\phi\in V_0$. We keep track of these \emph{proto-clusters} by defining 

\begin{equation*}
W_0 = \{ \phi \in V_0 \; \big\lvert \; \lvert P_\phi\lvert>1 \}\;.    
\end{equation*}

Now, we construct the reduced graph $G_0^{d_0}$. The clustering method searches for clusters by walking the reduced graph $G_0^{d_0}$, for some $d_0$, using a Breadth First Search (BFS) starting at the node $\varphi_0$. This BFS-walk enumerates the nodes in $V_0$ as a sequence $\varphi_0,\varphi_1,\dots,\varphi_N$. We group nodes together in sets $C_i\subset V_0$ if the set of points they represent has density greater or equal than a minimum density $\delta_0$, we call this the \emph{density condition}. We say that a set $C$ satisfies the density condition if 
\begin{equation}
\label{eq:density_cond}
    \text{diam}(C)\leq   \frac{\lvert C\lvert}{\delta_0} \;.
\end{equation}

\begin{algorithm}[htbp]\footnotesize
	\caption{HOS-Clustering}
	\label{alg:hos-c}
    \begin{algorithmic}[1]
    \Require {$A$: data set of points, $A=\{v_1,...,v_{N}\}\subset \mathbb{S}^{D-1} \subset \mathbb{R}^D$, with $v_i=\left(v_i^{(1)},...,v_i^{(D)} \right)$, $k_0$: minimum number of elements in a cluster ($k_0>1$), $\delta_0$: minimum density of every cluster,$d_0$: maximum Levenshtein distance between nodes of $G_0$.}
	    \State $R_{\Pi,\Theta}\gets\displaystyle\argmax_{R_{\Pi,\Theta}}\mathbf{c}(R_{\Pi,\Theta}A^T)$ 
	    \Comment{where $\mathbf{c}(A)=\frac{1}{N\sqrt{D}}\sum_{k=1}^N\sum_{j=1}^D \mid v_k^{(j)}\mid$}
		\State $A \gets R_{\Pi,\Theta} A^T$
		\State $V_0 \gets \{ h_v \; \mid \; v\in A\}$
		\State $P_{\phi} \gets \{ v \in A \; \mid \; \sigma_v \sim \phi \}$
		
		\State $W_0 \gets \{ \phi \in V_0 \; \mid \; \lvert P_\phi\rvert >1 \}$
		\State Build($G_0^{d_0}$)
		\State $\mathcal{C}_0\gets\{\}$ \Comment{empty collection of subsets of $V_0$}
		\State $\varphi_0 \gets \underset{\phi\in V_0}{\text{argmin}}  \sum_{\psi\in V_0} d_L(\phi,\psi)$
		\State $\mathbf{s}\gets\mathbf{BFS}(G_0^{d_0},\varphi_0)$ \Comment{This search enumerates the nodes in $V_0$ as a sequence $\mathbf{s} = \varphi_0,\dots,\varphi_N$}
		\State $i \gets 0,\; j\gets 0$
		\While{
		$j <= N$
		}
		  \State Form the possible cluster $C_i\gets\{\phi_j\}$
		  \While{$\textrm{diam}(C_i\cup\{\phi_{j+1}\})\leq\frac{\lvert C_i\rvert +1}{\delta_0} \;\mathbf{and}\; j<N$}
            \State add $\phi_{j+1}$ to $C_i$
            \State $j = j+1$
          \EndWhile
          \If{$\displaystyle \lvert\cup_{\phi\in C_i}P_\phi\rvert\geq k_0$}
            \State add the set $C_i$ to the collection $\mathcal{C}_0$
            \State $i = i+1$
          \EndIf
        \EndWhile
        \State $\displaystyle \mathcal{C}_1 \gets \mathcal{C}_0 \cup \Big\{ \{ \varphi \} \;\mid \; 
        \begin{array}{l}
            \varphi \in W_0, \\
            \varphi \not\in C\text{ for any } C\in  \mathcal{C}_0      
        \end{array}
        \Big\}$
        \State $\displaystyle \mathcal{C}_2 \gets  \Big\{ \bigcup_{\phi\in C} P_\phi \;\mid C\in  \mathcal{C}_1       \Big\}$
    \Ensure{$\mathcal{C}_{2}$}
\end{algorithmic}
\end{algorithm}

The result is a collection of subsets $C_i\subset V_0$, denoted by $\mathcal{C}_0$, where every $C_i$ must also satisfy the condition $\displaystyle \lvert \cup_{\phi\in C_i} P_\phi \lvert \geq k_0$, where $k_0$ is the minimum number of elements in a cluster. Out of the set of hyperparameters $\{d_0,k_0,\delta_0\}$, $\delta_0$ is actually the most important as it controls the resulting number of clusters. Next, we enrich this collection $\mathcal{C}_0$ with nodes from $W_0$ not belonging to any element of $\mathcal{C}_0$, and denote this enhanced collection by

\begin{equation*}
  \mathcal{C}_1 = \mathcal{C}_0 \cup \Big\{ \{ \varphi \} \;\Big\lvert \; 
    \begin{array}{l}
    \varphi \in W_0, \\
    \varphi \not\in C\text{ for any } C\in  \mathcal{C}_0      
    \end{array}
    \Big\}\;.  
\end{equation*}

Now, we take the subsets of $A$ corresponding to the elements of $\mathcal{C}_1$. We define

\begin{equation}
    \label{eq:clusters_c2}
    \mathcal{C}_2 = \Big\{ \bigcup_{\phi\in C} P_\phi \;\Big\lvert \; 
    C\in  \mathcal{C}_1      
    \Big\}\;.
\end{equation}
\noindent Observe that $\mathcal{C}_2$ is a collection of subsets of $A$. This is the collection of clusters. 

As a consequence of the density condition (\ref{eq:density_cond}), and (\ref{eq:clusters_c2}), our method produces a clustering of $A$ which is not a partition. We regard points not belonging to any cluster as \emph{noise}. The clustering method is summarized in Algorithm \ref{alg:hos-c}.

\begin{figure}[htbp]
\centering
\begin{subfigure}[t]{.3\linewidth}
    \scalebox{1}{
    \includegraphics[width=.7\linewidth, ]{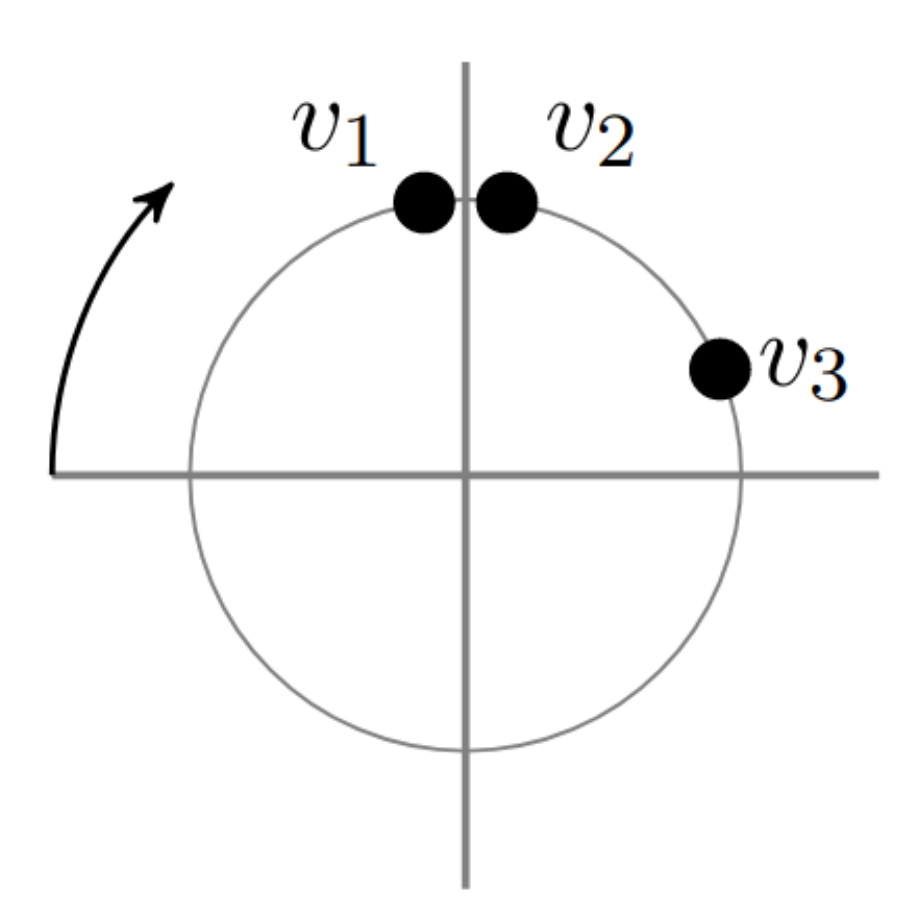}
    }
\caption{Apply rotation.}\label{fig:te1}
\end{subfigure}
\hspace{4em}
\begin{subfigure}[t]{.3\linewidth}
    \scalebox{0.9}{
    \includegraphics[width=\linewidth, ]{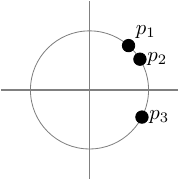}
    }
\caption{$V_0=\{++,+-\}$ and $W_0=\{++\}$.}\label{fig:te2}
\end{subfigure}

\begin{subfigure}[b]{.25\linewidth}
\centering
    \scalebox{1}{
    \includegraphics[width=.7\linewidth, ]{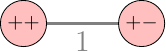}
    }
\caption{Construct graph $G_0$ and perform the BFS starting at $\phi_0=++$. }
\label{fig:te3}
\end{subfigure}
\hspace{4em}
\begin{subfigure}[b]{.25\linewidth}
\centering
    \scalebox{1.1}{
    \includegraphics[width=.7\linewidth, ]{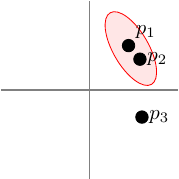}
    }
\caption{We find one cluster $C_1=\{p_1,p_2\}$.}
\label{fig:te4}
\end{subfigure}

\caption{Walk-through example illustrating the effect of rotation: $(a)$ A data set $A\subset \mathbb{S}^1\subset\mathbb{R}^2$ given by the three points, to which an adequate rotation $R$ is applied; $(b)$ Corresponding sets $V_0$ and $W_0$ after rotation; $(c)$ The graph $G_0$ and walk via BFS, using appropriate hyperparameters $\delta_0$ and $k_0>1$; $(d)$ Found cluster $C_2=\{p_1,p_2\}$.}
\label{fig:example_rot}
\end{figure}

To illustrate the clustering method we provide two walk-through examples depicted in Figures (\ref{fig:prin}) and (\ref{fig:example_rot}). It is worth stressing that the method works best with points in high-dimensional spaces. In low dimensional spaces, the method produces trivial results, except for very specific configurations of points; the examples are presented for illustrative purposes. The first example does not include rotations. The second example shows the effect of a rotation. In this example, if we choose a small enough $\delta_0$, we could end up with one cluster $C_2=\{p_1,p_2,p_3\}$. If we choose a large enough $\delta_0$, we could end up with no clusters.

\subsection{Experimental design to evaluate the method}
\label{subsec:topic-detection}
In this section, we describe the implementation of the clustering method as a tool for topic detection. 
The question we address in this work is how to systematically explore, via clustering, the similarities between text vector representations, in order to discover semantic relationships possibly encoded by patterns of signs in the embeddings. 

 We also describe the measures that we use for its evaluation, and its comparison with DBSCAN.

We start with a corpus of $N$ documents $\mathbf{d}_i$, each one with a label indicating the topic of the document denoted by an integer $y_i$. We will generate $D$-dimensional embeddings for the documents $\mathbf{d}_i$ using two main approaches: 

\begin{enumerate}
    \item Document embeddings using \texttt{doc2vec}.
    \item Means of word embeddings using \texttt{word2vec}, \texttt{FastText} and \texttt{GloVe}. We decided not to use transformer embeddings (for example, from BERT: \cite{devlin-etal-2019-bert}) because that implied designing evaluation measures adapted to the contextual representation of this type of models.
\end{enumerate}

In both approaches, each document $\mathbf{d}_i$ is represented by a point $v_i\in\mathbb{S}^{D-1}$. We apply the clustering method to the set of points $\{v_1,...,v_N\}$, yielding a set of clusters $\mathcal{C}=\{C_1,...,C_M\}$. We evaluate the quality of these clusters, both in terms of intrinsic coherence of the topic words in each cluster, and in terms of distance to the ground truth clustering. We use the following approaches:

\begin{enumerate}
    \item Topic Coherence using the cosine similarity \citep{newman2010automatic}. We define the topic coherence as follows. For each cluster $C\in\mathcal{C}$, we denote the set of documents represented by the cluster $C$ by $C^d = \{\mathbf{d}_{j_1},...,\mathbf{d}_{j_n}\}$. We take the $K$ most frequent words in $C^d$, and denote this set of words by $W_C^K$. Finally, given a word $w$, we denote by $\vv{w}$ a word embedding for $w$. The topic coherence of a cluster $C$ is given by

    $$\text{coh}(C)=\frac{\sum_{x,y\in W_C^K}\text{sim}(\vv{x},\vv{y})}{\binom{K}{2}},$$

    We define the topic coherence of the set of clusters $\mathcal{C}$ as

    $$\text{coh}\left(\mathcal{C}\right) = \frac{1}{\lvert \mathcal{C}\lvert }\sum_{C\in\mathcal{C}}\text{coh}(C).$$

    Observe that $\text{coh}\left(\mathcal{C}\right) \in \left[-1,1\right]$. The higher the value of $\text{coh}\left(\mathcal{C}\right)$ is, the more coherent are the documents in each cluster, topic-wise.
    
    \item Topic coherence using co-occurrences. This approach for evaluating a topic detection task is similar to the previous approach. However, instead of computing the semantic similarity between pair of topic words by means of the cosine similarity between their word embeddings, this approach computes their semantic similarity by counting co-occurrences using the pointwise mutual information \citep{islam2006second}.
    
    \item Topic Majority. We consider the clustering task as a classification task, assigning a predicted topic to each cluster based on the majority topic among the topics of each document in the cluster. In this way, we have a topic prediction for each document that can be compared with its actual topic. We calculate and report the accuracy of this classification task (acc). We also report how many clusters have a majority topic out of the total number of clusters (m/t).
    
    \item Adjusted Mutual Information. This is an information theoretic measure for clustering comparison quantifying the information shared by two clusters \citep{vinh2010information}. We use the implementation in \texttt{scikit-learn}. Here, it is worth pointing out that we are comparing both the baseline and our clustering method to the ground truth clustering, which might not be a clustering in the geometric sense. That is, documents sharing the same topic might not have vector representations close to each other. However, since we are comparing both clustering methods to the ground truth clustering, we report the results of this measure as a fair comparison score.
\end{enumerate}

\section{Experiments and Results}
\label{sec:experiments}

In this section we describe the experiments and results of the implementation of this clustering method in topic detection tasks using several text data sets. We compare this method with one of the most important density-based clustering methods, DBSCAN \citep{Ester1996DBSCAN}, in order to highlight its strengths and scope. In the next section we will interpret and discuss these results.

We use the following corpora of text documents:
\begin{itemize}
    \item \texttt{20NewsGroups}\footnote{\url{http://qwone.com/~jason/20Newsgroups/}}, which consists of 11000 documents. Each document in the corpus is marked with a single topic, there are 20 different topics.
    \item \texttt{BBC Full Text Document Classification}\footnote{\url{https://www.kaggle.com/shivamkushwaha/bbc-full-text-document-classification}}, which consists of 2225 documents from the BBC news website corresponding to stories in five topical areas from 2004-2005. There are 5 different topics: business, entertainment, politics, sport, tech.
    \item \texttt{Emotions data set for NLP}\footnote{\url{https://www.kaggle.com/praveengovi/emotions-data set-for-nlp}}, which is a collection of 16000 short documents, each document has an emotion flag. This is a data set used mainly for sentiment analysis, however, we regard the emotion of each document as the topic of the document.  
\end{itemize}

 
We used word and document embeddings with dimension $D=100$ in the three corpora and the four embeddings models. Our experiments were conducted using standard computer equipment Intel Core i7\textregistered, 2.80GHz $\times$ 8, and 16GB RAM.

We evaluate the clusters obtained using our method and DBSCAN using the four measures described in Section \ref{subsec:topic-detection}. The parameters for the clustering methods were chosen using a grid search, we use the parameters yielding the higher values of $\text{coh}$ whenever the resulting clustering is not trivial. 

In Tables \ref{tab:coherenceCOS_results} to \ref{tab:ami_clustering_results}, we report the scores of both clustering methods using the four measures.  

\begin{table}[!htbp]
\begin{center}
\caption{Coherence results, using the cosine similarity between word embeddings, for our clustering method compared to DBSCAN, in the three different corpora.}\label{tab:coherenceCOS_results}
\resizebox{0.75\textwidth}{!}{\begin{tabular}{l||c|c||c|c||c|c}
\toprule
\multicolumn{7}{c}{Topic coherence using the cosine similarity} \\
        \hline
        \multirow{2}{*}{Model}  & \multicolumn{2}{c||}{20NG} & \multicolumn{2}{c||}{BBC}  &\multicolumn{2}{c}{Emotion} \\
         & HOS-C & DBSCAN & HOS-C & DBSCAN & HOS-C & DBSCAN \\
        \hline
         \texttt{word2vec} & \textbf{0.753} &  0.751 & \textbf{0.847} & 0.734 & \textbf{0.895} & 0.796 \\
         \texttt{FastText} & \textbf{0.605} & 0.592 & \textbf{0.808} & 0.737 & \textbf{0.904} & 0.730 \\
          \texttt{GloVe} & \textbf{0.422} & 0.260 &  \textbf{0.392} & 0.256 &  \textbf{0.295} & 0.198 \\
          \texttt{doc2vec} & 0.397 & \textbf{0.491} & 0.512 & \textbf{0.581}  & \textbf{0.917} & 0.796 \\
\bottomrule
\end{tabular}}
\end{center}
\end{table}

\begin{table}[!htbp]
\begin{center}
\caption{Coherence results, using pointwise mutual information between topic words, for our clustering method compared to DBSCAN, in the three different corpora.}\label{tab:coherenceMI_results}
\resizebox{0.75\textwidth}{!}{\begin{tabular}{l||c|c||c|c||c|c}
\toprule
\multicolumn{7}{c}{Topic coherence using mutual information} \\
\hline
\multirow{2}{*}{Model}  & \multicolumn{2}{c||}{20NG} & \multicolumn{2}{c||}{BBC}  &\multicolumn{2}{c}{Emotion} \\
 & HOS-C & DBSCAN & HOS-C & DBSCAN & HOS-C & DBSCAN \\
\hline
 \texttt{word2vec} & \textbf{49.111} & 44.660 & \textbf{45.358}
 & 44.747 & \textbf{20.519} & 14.601 \\
 \texttt{FastText} & \textbf{37.899} & 37.573 & \textbf{47.858} & 41.103 & \textbf{12.001} & 4.811 \\
  \texttt{GloVe} & 36.132 & \textbf{72.123} & 42.860
  & \textbf{50.481} & \textbf{17.268} & 17.085 \\
  \texttt{doc2vec} & 36.029 & \textbf{72.495} & 34.049 & \textbf{41.349} & 22.502 & \textbf{26.065}\\

\bottomrule
\end{tabular}}
\end{center}
\end{table}

\begin{table}[!htbp]\footnotesize
\begin{center}
\caption{Accuracy results, using the topic majority classification, for our clustering method compared to DBSCAN, in the three different corpora.}\label{tab:accuracy_results}
\resizebox{\textwidth}{!}{\begin{tabular}{l|c|c|c|c|c|c|c|c|c|c|c|c}
\toprule
\multicolumn{13}{c}{Accuracy} \\
\hline
\multirow{3}{*}{Model}  & \multicolumn{4}{c}{20NG} & \multicolumn{4}{c}{BBC}  &\multicolumn{4}{c}{Emotion} \\
 & \multicolumn{2}{c}{HOS-C} & \multicolumn{2}{c}{DBSCAN} & \multicolumn{2}{c}{HOS-C} & \multicolumn{2}{c}{DBSCAN} & \multicolumn{2}{c}{HOS-C} & \multicolumn{2}{c}{DBSCAN} \\
 & m/t & acc & m/t & acc & m/t & acc & m/t & acc & m/t & acc & m/t & acc \\
 \hline
 \texttt{w2v} & 59/69 & \textbf{0.228} & 21/25 & 0.095 & 13/13 & \textbf{0.348} & 11/11 & 0.309 & 6/6 & 0.337 & 9/10 & 0.337\\
 \texttt{FT} & 59/75 & \textbf{0.263} & 31/34 & 0.137 & 6/6 & \textbf{0.405} & 18/20 & 0.325 & 3/3 & 0.336 & 14/15 & \textbf{0.339} \\
  \texttt{GV} & 3/5 & \textbf{0.253} & 14/15 & 0.198 & 3/3 & 0.234 & 5/5 & \textbf{0.775} & 56/60 & \textbf{0.399} & 17/18 & 0.338\\
  \texttt{d2v} & 1/1 & 0.200 & 5/5 & \textbf{0.215} & 5/5 & 0.555 & 8/9 & \textbf{0.448} & 40/41 & \textbf{0.369} & 5/6 & 0.335 \\

\bottomrule
\end{tabular}}
\end{center}

\end{table}

\begin{table}[!htbp]
\begin{center}
\caption{Adjusted Mutual Information results, both our clustering method and DBSCAN are compared against the topic labels, in the three different corpora. }\label{tab:ami_clustering_results}
\resizebox{0.75\textwidth}{!}{\begin{tabular}{l||c|c||c|c||c|c}
\toprule
\multicolumn{7}{c}{Clustering adjusted mutual information} \\
\hline
\multirow{2}{*}{Model}  & \multicolumn{2}{c||}{20NG} & \multicolumn{2}{c||}{BBC}  &\multicolumn{2}{c}{Emotion} \\
 & HOS-C & DBSCAN & HOS-C & DBSCAN & HOS-C & DBSCAN \\
\hline
 \texttt{word2vec} & \textbf{0.166} & 0.024 & \textbf{0.111} & 0.047 & \textbf{0.004} & 0.001 \\
 \texttt{FastText} & \textbf{0.123} & 0.036 & \textbf{0.135} & 0.062 & 0.001 & 0.001 \\
  \texttt{GloVe} & 0.010 & \textbf{0.034} & 0.021 & \textbf{0.209} & \textbf{0.004} & 0.001	\\
  \texttt{doc2vec} & 0.001 & \textbf{0.024} &  0.040 & \textbf{0.096} & 0.001 & 0.001 \\

\bottomrule
\end{tabular}}
\end{center}
\end{table}

\section{Discussion}
\label{sec:discussion}

As we have previously set out, we evaluated and compared our clustering method using three types of measures. 
Firstly, we use coherence-based measures (Tables \ref{tab:coherenceCOS_results}, \ref{tab:coherenceMI_results}), these measures account for how coherent are the most frequent words in the set of documents represented in each cluster. These measures do not take into account the original topics of the documents, instead they intrinsically evaluate the coherence of the topics in each cluster. Table \ref{tab:coherenceCOS_results} shows that our method performs better than DBSCAN. On the other hand, Table \ref{tab:coherenceMI_results} shows that our method has a similar performance to DBSCAN.

\begin{figure}[H]
\centering
  \begin{subfigure}[t]{0.32\linewidth} 
    \includegraphics[width=\linewidth]{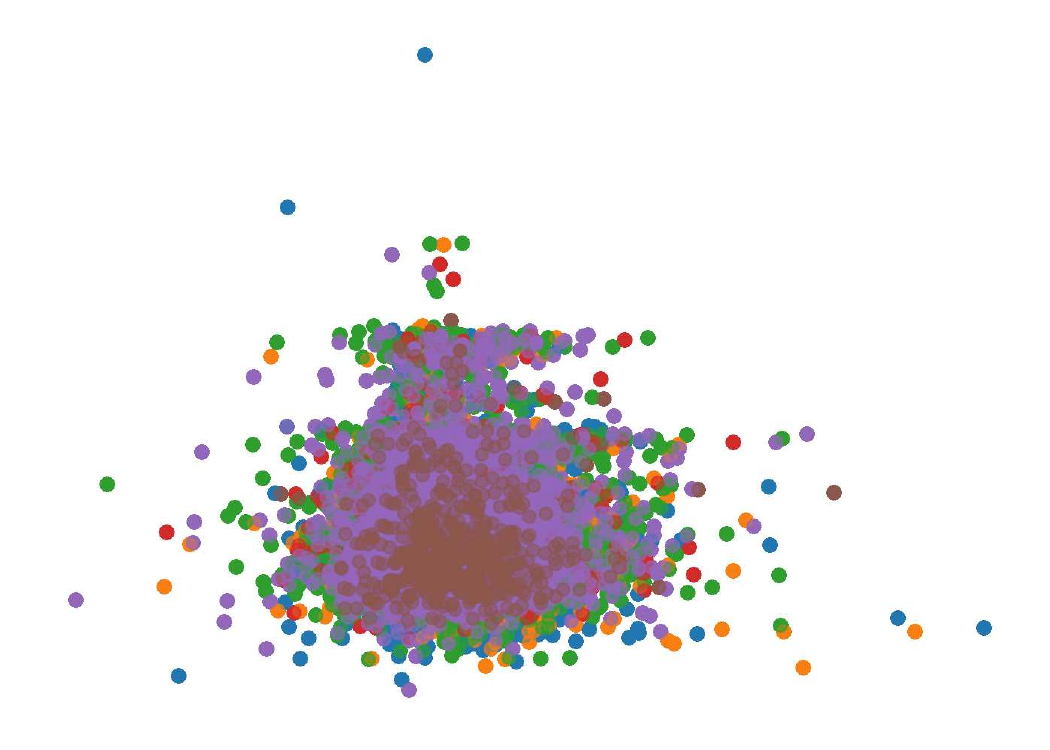} 
    \caption{Topics of the documents.}
    \label{fig:clusters_truth}
  \end{subfigure}
  \hfill 
  \begin{subfigure}[t]{0.32\linewidth}
    \includegraphics[width=\linewidth]{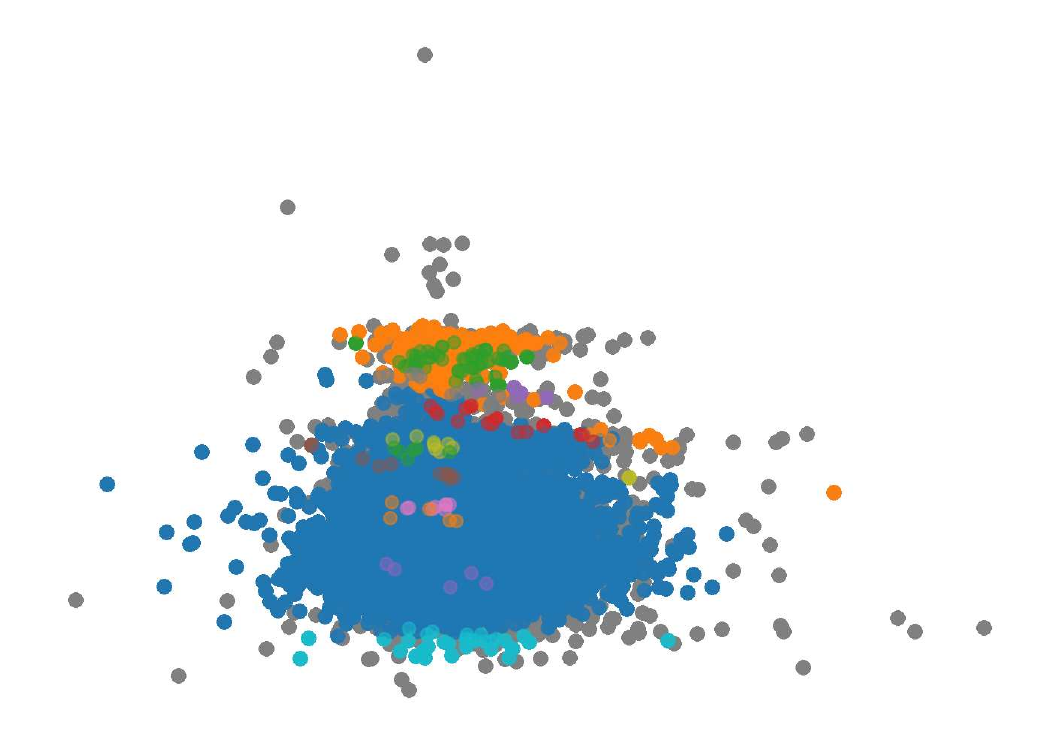}
    \caption{Clusters obtained with DBSCAN.}
    \label{fig:clusters_DBSCAN}
  \end{subfigure}
  \hfill
  \begin{subfigure}[t]{0.32\linewidth}
    \includegraphics[width=\linewidth]{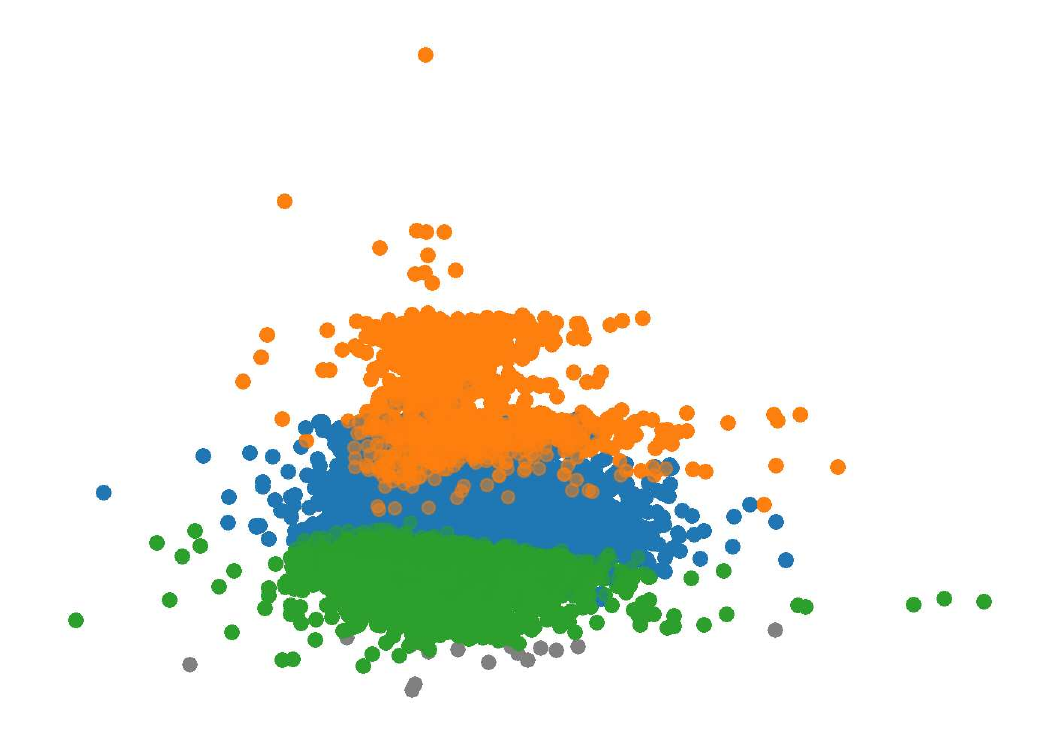}
    \caption{Clusters obtained with HOS-clustering.}
    \label{fig:clusters_ho}
  \end{subfigure}
\caption{Dimensionality reduction, using \texttt{PCA}, after clustering of the \texttt{Emotion} data set with \texttt{FastText} embeddings. Although the actual behaviour of the points in $\mathbb{R}^D$ may be different, it is very likely that there is a high overlap between the different topics, which makes clustering difficult.  } 
\label{fig:clusters_pca}
\end{figure}

Secondly, we measure the accuracy of the classification task (Table \ref{tab:accuracy_results}). As we can observe, our method performs better than DBSCAN. Lastly, we use adjusted mutual information, which is an information theoretic measure for clustering comparison (Table \ref{tab:ami_clustering_results}). As we have previously stated, we are comparing both clustering methods to the ground truth clustering, which might not be a good clustering in the geometric sense (we conjecture this from evidence depicted in Figure \ref{fig:clusters_pca}). As a consequence of this, the values of Table \ref{tab:ami_clustering_results} are low for both methods. This evidence shows that both methods have similar performances when evaluated with this measure. Other information theoretic measure, such as mutual information and adjusted rank index, exhibited similar scores. 

As a side remark, it is worth noting that both clustering methods performed systematically worse on the data sets consisting of \texttt{GloVe} word embeddings. Although there are techniques aiming to normalize the values of entropy-based measures for clustering comparison, so that they show comparable values across different data sets \citep{luo2008information}, we decided to keep the original values to show this interesting phenomenon. We will study this phenomenon in future work to shed some light into the difference of the four word embeddings models when capturing semantic regularities regarding topic detection. 

Taking into account all evaluation metrics on both clustering methods, we conclude that our clustering method allows to generate text clusters having a quality similar to the state of the art. However, our method shows advantages and differences over DBSCAN, which we now enumerate.

\begin{figure}[H]  
\centering 
  \begin{subfigure}[t]{0.46\linewidth}
    \centering
    \includegraphics[width=0.8\linewidth, , ]{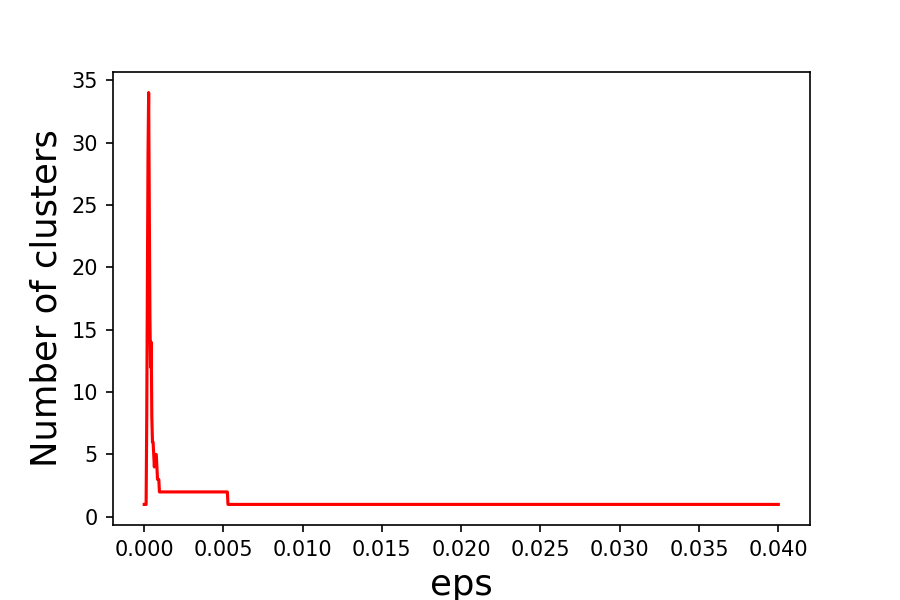}
    \caption{DBSCAN applied to the \texttt{word2vec} embeddings of the \texttt{BBC} data set.}
    \label{fig:DBSCAN_pdf_numclusters_1}
  \end{subfigure}
  \begin{subfigure}[t]{0.46\linewidth}
    \centering
    \includegraphics[width=0.8\linewidth, , ]{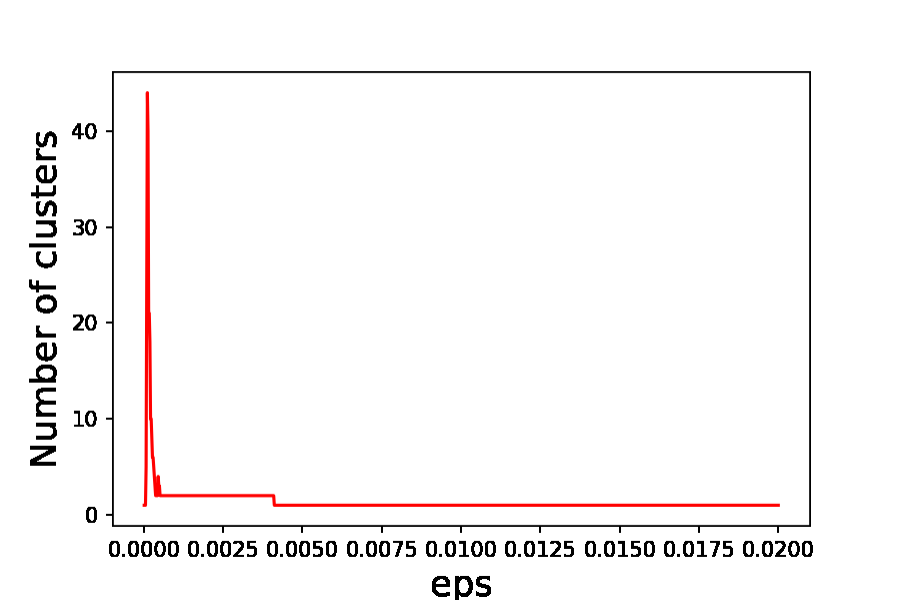}
    \caption{DBSCAN applied to the \texttt{FastText} embeddings of the \texttt{Emotion} data set.}
    \label{fig:DBSCAN_pdf_numclusters_2}
\end{subfigure}

  \begin{subfigure}[t]{0.46\linewidth}
    \centering
    \includegraphics[width=0.8\linewidth, , ]{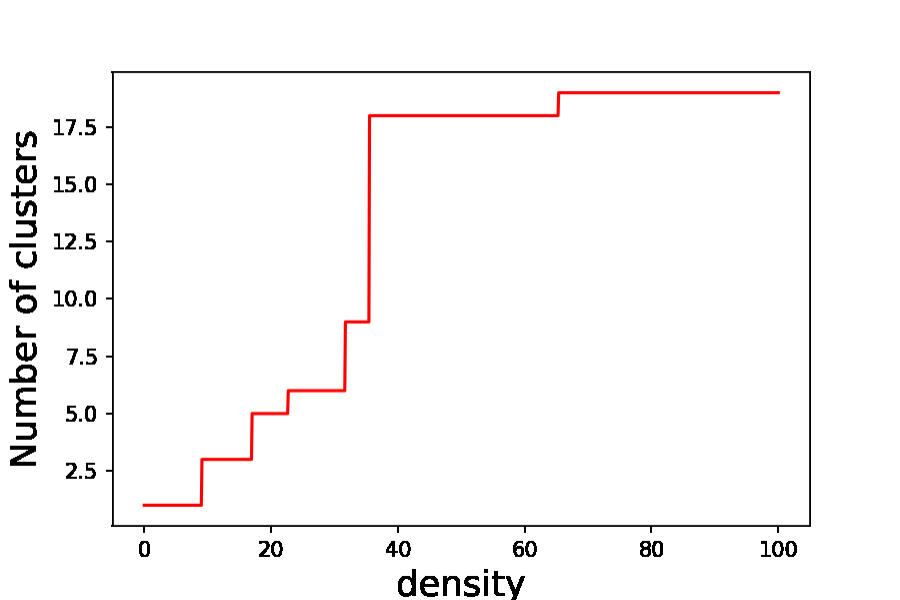}
    \caption{HOS-Clustering applied to the \texttt{word2vec} embeddings of the \texttt{BBC} data set.}
    \label{fig:ho_den_numclusters_1}
  \end{subfigure}
  \begin{subfigure}[t]{0.46\linewidth}
    \centering
    \includegraphics[width=0.8\linewidth, , ]{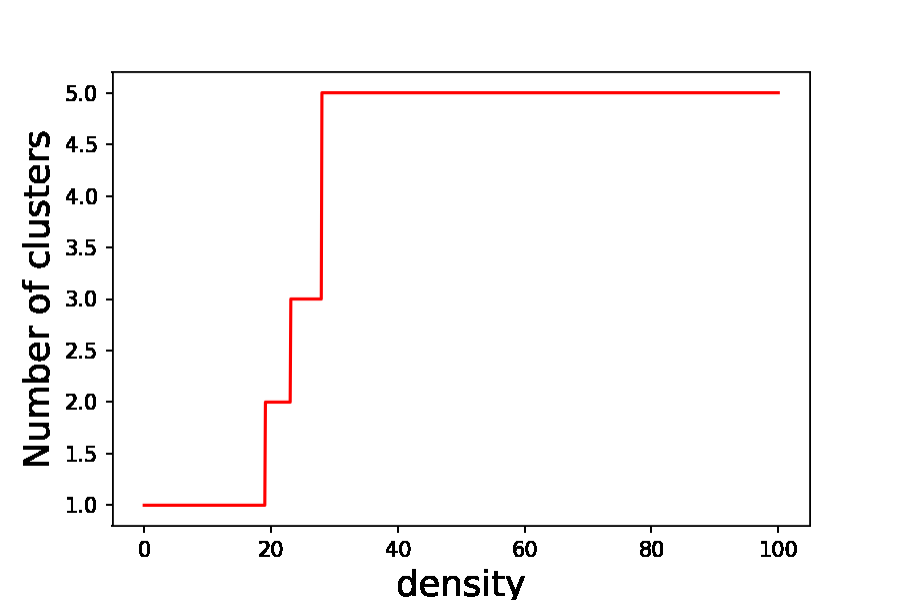}
    \caption{HOS-Clustering applied to the \texttt{FastText} embeddings of the \texttt{BBC} data set.}
    \label{fig:ho_den_numclusters_2}
\end{subfigure}

\caption{Number of clusters found by DBSCAN as a function of the hyperparameter $\varepsilon$ (top); and number of clusters found by HOS-Clustering as a function of the hyperparameter $\delta_0$ (bottom).}
\label{fig:DBSCAN_num_clusters_vs_pdf}
\end{figure}

DBSCAN strongly depends on the hyper parameter $\varepsilon$, which accounts for the maximum distance between two samples, for one to be considered in the vicinity of the other. Appropriately choosing this parameter is not a trivial task. Some heuristics for choosing this parameter have been discussed in the literature \citep{schubert2017DBSCAN}. However, we found that in this high-dimensional case, a small variation of $\varepsilon$ produces large variations in the number of clusters, and  overall in the clustering results. Moreover, the range of $\varepsilon$ values that give non-trivial clustering results is very narrow. Examples of this are shown in Figures \ref{fig:DBSCAN_pdf_numclusters_1} and \ref{fig:DBSCAN_pdf_numclusters_2}.

In contrast, our evidence suggests that our method is more stable under variations of the main hyper-parameter $\delta_0$. If $\delta_0$ is small, all points in hyperoctants represented by the nodes of the graph $G_0^{d_0}$ will form one cluster. On the other hand, if $\delta_0$ is large, we will have as many clusters as \emph{proto-clusters} $P\in W_0$ satisfying that $|P|>k_0$. This number is the maximum \emph{resolution} of our method. This simplifies the task of fine tuning the method in order to produce better results, since the interval of values of $\delta_0$ yielding non-trivial clustering results has infinite length. This behaviour is depicted in Figures \ref{fig:ho_den_numclusters_1}, \ref{fig:ho_den_numclusters_2}. 
This previously described scenario explains why our method is only able to find at most 18 clusters in \ref{fig:ho_den_numclusters_1}, and 5 in \ref{fig:ho_den_numclusters_2}.

Finally, but not the least, it is worth noting that our method is not only a clustering method, but also a tool for exploring the data set from a topological perspective. The reduction in the size of the set of entities when we go from $A$ to $V_0$ gives information about how concentrated are the points in the space. In other words, we know in how many regions the data set is located. For example, in the \texttt{BBC} corpus, using the \texttt{FastText} embeddings, after the appropriate rotation, there are only 16 hyperoctants containing the 2225 points. On the other hand, using the \texttt{GloVe} embeddings, there are 2070 hyperoctants containing the 2225 points. This difference immediately indicates that \texttt{GloVe} embeddings are more sparsely distributed in space than \texttt{FastText} embeddings.

\begin{figure}[H]
    \centering
    \begin{subfigure}[t]{\linewidth}
    \centering
    \includegraphics[width=0.8\linewidth, , ]{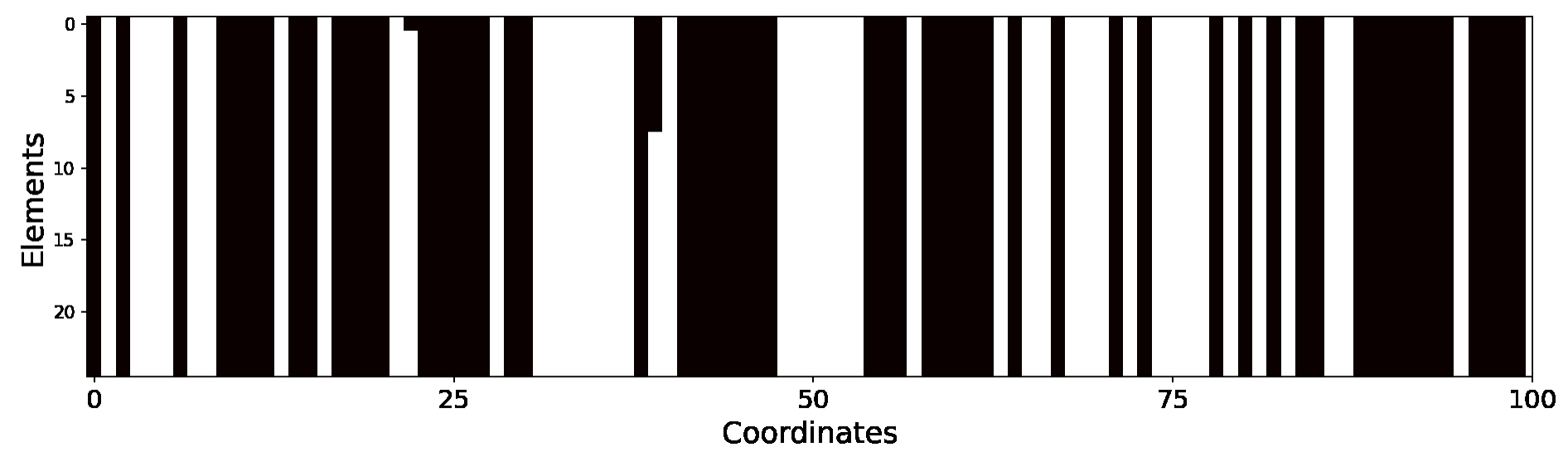}
    \caption{Signs of the coordinates of each document embedding in a cluster.}
    \label{fig:bar_cluster}
    \end{subfigure}
    
    \begin{subfigure}[t]{\linewidth}
    \centering
    \includegraphics[width=0.8\linewidth, , ]{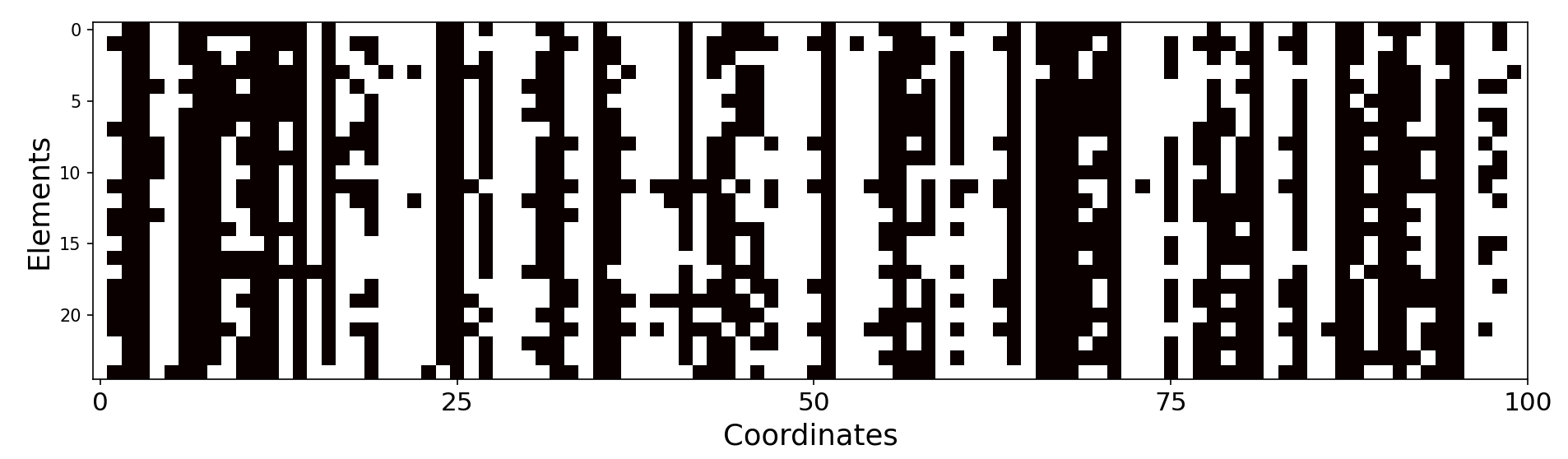}
    \caption{Signs of the coordinates of each document embedding in a random set of embeddings.}
    \label{fig:bar_non_cluster}
    \end{subfigure}
    \caption{Two different sets of 25 document \texttt{FastText} embeddings in the \texttt{Emotion} corpus. Each plot is a rectangular $25\times 100$ grid, where each row represents the signs of the 100 coordinates of the embedding  ($+$:black; $-$: white) : a) A set of embeddings displaying only three different combinations of signs in their coordinates, thus evidencing the fact that these 25 embeddings forms one cluster; b) A set of embeddings displaying many different different combinations of signs in their coordinates, thus evidencing the fact that they do not form any cluster.}
    \label{fig:cluster-bar-plot}
\end{figure}

For illustrative purposes of the effects of clustering under our approach, in Figure \ref{fig:cluster-bar-plot} we show the difference in sign vectors between a cluster of points (Figure \ref{fig:bar_cluster}, and a random set of points (Figure \ref{fig:bar_non_cluster}). Each plot represents a set $A$ of $n$ $d$-dimensional vectors as a heatmap of a $n\times d$ matrix with entries $\{-1,1\}$. Each row is the sign vector of a vector of $A$. In this case, $n=30$, $d=100$. Both sets are subsets of the \texttt{FastText} documents embeddings of the \texttt{Emotion} data set. The difference between these two graphs provides insight into clustering using hyperoctants. As we observe, in the cluster depicted in Figure \ref{fig:bar_cluster} there are only points in three different hyperoctants.

Now, we discuss the weaknesses of our method. Consider the following scenario in a data set $A$, there are two subsets $C_1,C_2 \subset A$ which are ground truth clusters, such that $\text{diam}(C_1\cup C_2)<\frac{\pi}{2}$. In other words, we have two ground truth clusters that, after an appropriate rotation, lie in the same hyperoctant. In this scenario, our method will join the points in $C_1\cup C_2$ in only one cluster. However, as we discussed in Observation \ref{obs:points_same_HO}, as the dimension increases, the probability of such a situation occurring decreases exponentially, and the fact that two points are in the same hyperoctant becomes more meaningful in terms of clustering. In future work, we will address this limitation by properly embedding such hyperoctants in a higher dimensional space. 

Moreover, as we have described, our method is suitable for data sets in spaces endowed with the angular metric, which could be seen as a limitation. However, if an Euclidean $D$-space is properly embedded in a $D$-dimensional $\mathbb{R}^{D+1}$-sphere, it is possible to apply our clustering method to this embedded data set. This will be described in a future paper.

Regarding the question of how big should be the dimension $D$ of the feature space in order to use our method, it should satisfy 

\begin{equation}\label{eq:condition_dimension}
    D>\frac{\log N}{\log 2}
\end{equation}
    
where $N$ is the number of points in the data set. This guarantees that there is enough hyperoctants to contain each point of the data set. Thus, giving significance to the fact that two points lie in the same hyperoctant. If condition (\ref{eq:condition_dimension}) does not hold, at least two points are bound to lie in the same hyperoctant, not because they are necessarily close to each other, but because of the pigeonhole principle. Addressing the behaviour of this method in even higher dimensions, we would like to mention here that we conducted experiments with these corpora using BERT's vector representations (with dimension 512), but they were not reported because the coherence metrics, used for the exhaustive comparison, are not adapted to the properties of these embeddings. This will be part of future work.

As a last point of this section, we would like to briefly discuss the connections our clustering method has with other research fields. A first aspect is the connections that the graph of hyperoctants $P_D$ suggests. $P_D$ can be regarded as the Hasse diagram of the power-set lattice $P_{\text{fin}}(\phi(A))$ \citep{davey2002introduction}. It is also isomorphic to the binary Hamming space, which is the set of all $2^N$ binary strings of length N, used in the theory of signals coding and transmission \citep{baylis2018error}. 

Furthermore, there are other possible applications and connections of our approach to code theory. Using the language of this research realm \citep{lint1995coding}, and our notation, we rotate our original data set $A$, thus obtaining the data set $V_0$ which will be clustered. This rotated data set is a $(D,N,\alpha)$ code of length $N$ in the space $\mathbb{F}_2^D$. This code has certain minimum distance $\alpha\in[0,D]$, and some covering radius $r\in[0,D]$. Here, $\mathbb{F}_2$ denotes the finite field of two elements. This code is not necessarily linear. 

To briefly illustrate our point, we explore two concepts of code theory, applied to our setting. Firstly, we explore the covering radius $r$. A $r$-covering binary code is a finite subset $V\subset \mathbb{F}_2^D$ such that, for every code-word $y\in\mathbb{F}_2^D$, there exist $x\in V$ such that $d_L(x,y)\leq r$. In other words, the union of all closed balls of radius $r$ around code-words of $V$ cover the space $\mathbb{F}_2^D$. The covering radius of the code $V$ is the smallest $r$ such that $V$ is $r$-covering. Covering codes is a central object in coding theory, they have found applications in diverse areas. Generally, one is interested in the estimation of $K(D,r)$, the minimum cardinality of a binary code of length $D$ and covering radius $r$. 
For a data set $V_0$, the bigger the covering radius, the more evenly distributed are the points in the different hyperoctants, and thus, the least discernible are the possible clusters. A clear cluster should be a $r$-covering sub-code $B\subset V_0$ with a large $r$. Thus, the clustering method translates into finding different subsets $C\subset V_0$ such that every one of these sets $C$ is a $r$-covering code with a large $r$ depending on the value of $\delta_0$. In this setting, given a minimum density $\delta_0$, we can estimate $d_0$, and therefore, $K(D,d_0)$ gives a lower bound for the number of instances necessary for obtaining a non-trivial clustering. 
On the other hand, linearity imposes strong restrictions on $V_0$, and ultimately, on the possible coordinates of the points of the data set. For example, for every two code-words $x,y\in V_0$, it should hold $x+y\in V_0$. Linear codes exhibit regularities in $\mathbb{F}_2^D$, which a general data set might not satisfy. Therefore, in the setting of this method, we are interested in non-linear codes.

These previous arguments show the potential interplay between our clustering method and code theory. These ideas will be developed in a future paper.

A last word on the applications of the proposed method is worth mentioning. Our method could also be useful in the analysis of high-dimensional data such as images or genomic sequences \citep{campo2020convex}. Finally, since there are many contributions using the Hamming and Levenshtein distances, or sign matrices, our method could have an impact on other realms of research such as bioinformatics \cite{logan_3gold_2022, campo2020convex, wang_using_2015}, pattern recognition \cite{fan_efficient_2020}, image retrieval \cite{fakhfakh_image_2015}, data mining \cite{Zhang2006DM}, communication complexity and learning theory \cite{lokam2009complexity}.

\section{Conclusions}
\label{sec:conc}
Clustering is part of the exploratory data methods that aim at enhancing the practitioner’s natural ability to recognize patterns in the data being studied. 

In this work, we proposed a new clustering method called Hyperoctant-Search Clustering, based on a combinatorial-topological approach applied to regions of space defined by signs of coordinates (hyperoctants). The method builds clusters of data points based on the partitioning of a connected graph, whose vertices represent hyperoctants, and whose edges connect neighboring hyperoctants under the Levenshtein distance. 

Clustering has many useful applications. As our method is best suited for high-dimensional spaces, where data points contain positive and negative features ---typical examples of which are dense vector representations (embeddings) produced by neural networks--- in this paper we selected topic detection to assess the performance of the proposed method.
Our objective was to explore, via clustering, the similarities between dense vector representations of documents, in order to discover semantic relationships possibly encoded by patterns of signs in the embeddings. 

Our experimental results provide evidence that our method is a sound alternative to perform clustering tasks in high-dimensional spaces endowed with the angular metric, allowing also to acquire insight about some topological properties of the semantic space, and having a stable performance under variations of its most important hyperparameter.

The ability provided by the proposed method to explore data from a topological perspective extends the current possibilities of exploring high-dimensional data. As our experiments show, the method helps to gain insight into the distribution of the embeddings in the semantic space, pinpointing both void and highly populated regions, as it directly provides information on the number of hyperoctants where there are data points. The smaller this number is compared to the total number of points, the more ``irregular'' the distribution of the set of points in space is. In this sense, our method works better the more ``irregularities'' there are in the distribution of the set of points in space; that is, the more clusterable the input problem is. This phenomenon, intrinsic to the clustering problem, has already been extensively discussed in the past, where it has been observed that if the problem is not clusterable from the start, no method will be able to solve it. As  \cite{ackerman09a} have eloquently set out: \textit{“if a data set is hard to cluster then it is does not have a meaningful clustering structure”, [...] the hard cases that render clustering NP-hard are the inputs we don’t care to cluster}\citep[p. 1]{ackerman09a}. 
An in-depth study of this topological data analysis capability of our method will be part of future work. 

\subsection*{Acknowledgments}
The authors acknowledge the support of CONAHCYT (now SECIHTI).


\bibliographystyle{unsrt}
\bibliography{references}

\end{document}